\documentclass[10pt,letterpaper]{article}

\usepackage{cvpr}
\usepackage{times}
\usepackage{epsfig}
\usepackage{graphicx}
\usepackage{soul}
\usepackage[export]{adjustbox}
\usepackage{amsmath}
\usepackage{amssymb}
\usepackage{mathtools}
\newtheorem{theorem1}{Theorem}

\usepackage{comment}
\usepackage{bm}
\usepackage{mdframed}
\usepackage{authblk}

\usepackage{lipsum}

\newcommand\blfootnote[1]{%
  \begingroup
  \renewcommand\thefootnote{}\footnote{#1}%
  \addtocounter{footnote}{-1}%
  \endgroup
}

\usepackage{hyperref}
\usepackage{caption}
\usepackage{subcaption}

\cvprfinalcopy 

\newcommand{\tblcaption}[1]{\def\@captype{table}\caption{#1}}

\def\cvprPaperID{4645} 

\ifcvprfinal\fi
\begin{document}

\setlength{\abovedisplayskip}{2pt}
\setlength{\belowdisplayskip}{2pt}

\title{Unsupervised Visual Domain Adaptation:\\
A Deep Max-Margin Gaussian Process Approach}

\author[1,2]{Minyoung Kim\thanks{mikim21@gmail.com}}
\author[1]{Pritish Sahu\thanks{ps851@cs.rutgers.edu}}
\author[1]{Behnam Gholami\thanks{bb510@cs.rutgers.edu}}
\author[1]{Vladimir Pavlovic\thanks{vladimir@cs.rutgers.edu}}
\affil[1]{Dept. of Computer Science, Rutgers University, NJ, USA}
\affil[2]{Dept. of Electronic Engineering, Seoul National University of Science $\&$ Technology, South Korea}


\maketitle

\blfootnote{Pritish Sahu and Behnam Gholami contributed equally to this work.}

\begin{abstract}
In unsupervised domain adaptation, it is widely known that the target domain error can be provably reduced by having a shared input representation that makes the source and target domains indistinguishable from each other. Very recently it has been studied that not just matching the marginal input distributions, but the alignment of output (class) distributions is also critical. The latter can be achieved by minimizing the maximum discrepancy of predictors (classifiers). In this paper, we adopt this principle, but propose a more systematic and effective way to achieve hypothesis consistency via Gaussian processes (\textbf{GP}). The \textbf{GP} allows us to define/induce a hypothesis space of the classifiers from the posterior distribution of the latent random functions, turning the learning into a simple large-margin posterior separation problem, far easier to solve than previous approaches based on adversarial minimax optimization. We formulate a learning objective that effectively pushes the posterior to minimize the maximum discrepancy. This is further shown to be equivalent to maximizing margins and minimizing uncertainty of the class predictions in the target domain, a well-established principle in classical (semi-)supervised learning. Empirical results demonstrate that our approach is comparable or superior to the existing methods on several benchmark domain adaptation datasets.
\end{abstract}
\section{Introduction}\label{sec:intro}

The success of deep visual learning largely relies on the abundance of data annotated with ground-truth labels where the main assumption is that the training and test data follow from the same underlying distribution. However, in real-world problems this presumption rarely holds due to a number of artifacts, such as the different types of noise or sensors, changes in object view or context, resulting in degradation of performance during inference on test data.  One way to address this problem would be to collect labeled data in the test domain and learn a test-specific classifier while possibly leveraging the model estimated from the training data.  Nevertheless, this would typically be a highly costly effort.

Domain adaptation, a formalism to circumvent the aforementioned problem, is the task of adapting a model trained in one domain, called the {\em source}, to another {\em target} domain, where the source domain data is typically fully labeled but we only have access to images from the target domain with no (or very few) labels. Although there are several slightly different setups for the problem, in this paper we focus on the {\em unsupervised domain adaptation} (\textbf{UDA}) with {\em classification} of instances as the ultimate objective.  That is, given the fully labeled data from the source domain and unlabeled data from the target, the goal is to learn a classifier that performs well on the target domain itself. 


One mainstream direction to tackle \textbf{UDA} is the shared space embedding process. The idea is to find a latent space shared by both domains such that the classifier learned on it using the fully labeled data from the source will also perform well on the target domain. This is accomplished, and supported in theory~\cite{ben-david-2007}, by enforcing a requirement that the distributions of latent points in the two domains be indistinguishable from each other. 
A large family of \textbf{UDA} approaches including~\cite{gopalan2011domain,gong2013connecting,baktashmotlagh2013unsupervised,ganin2014unsupervised,long2014transfer,kan2015bi,gholami2017punda,ming2015unsupervised,ghifary2016deep} leverage this idea (see Sec.~\ref{sec:related} for more details). However, their performance remains unsatisfactory, in part because the methods inherently rely on matching of marginal, class-free, distributions while using the underlying assumption that the shift in the two distributions, termed covariate shift \cite{sugiyama2008direct}, can be reduced without using the target domain labels. 



To address this issue, an effective solution was proposed in~\cite{saito2018}, which aims to take into account the class-specific decision boundary. Its motivation follows the theorem in~\cite{ben-david-2010} relating the target domain error to the maximal disagreement between any two classifiers, tighter than the former bound in~\cite{ben-david-2007}. It implies that a provably small target error is achievable by minimizing the maximum classifier discrepancy (\textbf{MCD}). The approach in~\cite{saito2018}, the \textbf{MCD} Algorithm (\textbf{MCDA} for short), attempted to minimize \textbf{MCD} directly using adversarial learning similar to GAN training~\cite{gan14}, i.e., through solving a minimax problem that finds the pair of most discrepant classifiers and reduces their disagreement.  

In this paper we further extend the \textbf{MCD} principle 
by proposing a more systematic and effective way to achieve consistency in the hypothesis space of classifiers $\mathcal{H}$ through Gaussian process (\textbf {GP})~\cite{gpml_book} endowed priors, with deep neural networks (DNNs) used to induce their mean and covariance functions. The crux of our approach is to regard the classifiers as random functions and use their posterior distribution conditioned on the source samples, as the prior on $\mathcal{H}$.  The key consequence and advantages of this Bayesian treatment are: (1) One can effectively minimize the inconsistency in $\mathcal{H}$ over the target domain by regularizing the source-induced prior using a max-margin learning principle~\cite{wang2013max}, a significantly easier-to-solve task than the minimax optimization of~\cite{saito2018} which may suffer from the difficulty of attaining an equilibrium point coupled with the need for proper initialization. 
(2) We can quantify the measure of prediction uncertainty and use it to credibly gauge the quality of prediction at test time.
 
Although \textbf {GP} models were previously known to suffer from the scalability issues~\cite{gpml_book}, we utilize recent deep kernel techniques~\cite{deep_kernel,dkl16} to turn the non-parametric Bayesian inference into a more tractable parametric one, leading to a learning algorithm computationally as scalable and efficient as conventional (non-Bayesian) deep models.
%
Our extensive experimental results on several standard benchmarks demonstrate that the proposed approach achieves state-of-the-art prediction performance, outpacing recent \textbf{UDA} methods including \textbf{MCDA}~\cite{saito2018}.

\section{Problem Setup and Preliminaries}\label{sec:setup}

We begin with the formal description of the \textbf{UDA} task for a multi-class classification problem. 

\textbf{Unsupervised domain adaptation:} {\em Consider the joint space of inputs and class labels, $\mathcal{X}\times\mathcal{Y}$ where 
$\mathcal{Y}=\{1,\dots,K\}$ for ($K$-way) classification. Suppose we have two domains on this joint space, \textbf{source (S)} and \textbf{target (T)}, defined by unknown distributions $p_S({\bf x},y)$ and $p_T({\bf x},y)$, respectively.  We are given source-domain training examples with labels $\mathcal{D}_S = \{({\bf x}^S_i, y^S_i)\}_{i=1}^{N_S}$ and target data $\mathcal{D}_T = \{{\bf x}^T_i\}_{i=1}^{N_T}$ with no labels.  We assume the shared set of class labels between the two domains. The goal is to assign the correct class labels $\{y^T_i\}$ to target data points $\mathcal{D}_T$.}

To tackle the problem in the shared latent space framework, we seek to learn the embedding function ${\bf G}:\mathcal{X}\to\mathcal{Z}$ and a classifier $h:\mathcal{Z}\to\mathcal{Y}$ in the \textit{shared latent space} $\mathcal{Z}$. The embedding function ${\bf G}(\cdot)$ and the classifier $h(\cdot)$ are shared across both domains and will be applied to classify samples in the target domain using the composition $y = h({\bf z}) = h({\bf G}({\bf x}))$.  

Our goal is to find the pair $(h,{\bf G})$ resulting in the lowest generalization error on the target domain, 
\vspace{+0.5em}
\begin{equation}
(h^*,{\bf G}^*) = \arg\min_{h,{\bf G}} e_T(h,{\bf G}) 
  = \arg\min_{h,{\bf G}} \mathbb{E}_{({\bf x},y)\sim p_T({\bf x},y)}[I(h({\bf G}({\bf x})) \neq y)],
\vspace{+0.3em}
\end{equation}
with $I(\cdot)$ the $1/0$ indicator function.  Optimizing $e_T$ directly is typically infeasible.  Instead, one can exploit the upper bounds proposed in~\cite{ben-david-2010} and~\cite{ben-david-2007}, which we restate, without loss of generality, for the case of fixed ${\bf G}$.

\begin{theorem1}\label{theorem}\cite{ben-david-2010,ben-david-2007}
Suppose that $\mathcal{H}$ is symmetric (i.e.,  $h\in\mathcal{H}$ implies $-h\in\mathcal{H}$). For any $h\in\mathcal{H}$, the following holds\footnote{Note that the theorems assume binary classification ($y\in \{+1,-1\}$), however, they can be straightforwardly extended to multi-class setups.}:
\vspace{+0.3em}
\begin{equation}
e_T(h) \ \leq \  
  e_S(h) + \sup_{h,h' \in \mathcal{H}} 
    \big\vert d_S(h,h') - d_T(h,h') \big\vert  + e^* \label{eq:thm_2010}
\end{equation}
\begin{equation}
\ \ \ \ \ \ \ \ \ \ \ \leq \  
  e_S(h) + \sup_{h \in \mathcal{H}} \big\vert d_S(h,+1) - d_T(h,+1) ] \big\vert  + e^* 
  \label{eq:thm_2007}
\end{equation}
\end{theorem1}
Here $e_S(h)$ is the error rate of $h(\cdot)$ on the source domain, $e^*:= \min_{h\in\mathcal{H}} e_S(h) + e_T(h)$, and $d_S(h,h') := \mathbb{E}_{{\bf z}\sim S} [ \mathbb{I}(h({\bf z}) \neq h'({\bf z})) ]$ denotes the discrepancy between two classifiers $h$ and $h'$ on the source domain $S$, and similarly for $d_T(h,h')$.  We use ${\bf z}\sim S$ to denote the distribution of ${\bf z}$ in the latent space induced by ${\bf G}$ and $p_S({\bf x},y)$.

\textbf{Looser bound.} With $e^*$ the uncontrollable quantity, due to the lack of labels for $T$ in the training data, the optimal $h$ can be sought through minimization of the source error $e_S(h)$ and the worst-case discrepancy terms.  In the looser bound (\ref{eq:thm_2007}), the supremum term is, up to a constant, equivalent to $\sup_{h \in \mathcal{H}} \mathbb{E}_{{\bf z}\sim S} [ I(h({\bf z}) = +1) ] + \mathbb{E}_{{\bf z}\sim T} [ I(h({\bf z}) = -1) ]$, the maximal accuracy of a domain discriminator (labeling $S$ as $+1$ and $T$ as $-1$). Hence, to reduce the upper bound one needs to choose the embedding ${\bf G}$ where the source and the target inputs are indistinguishable from each other in $\mathcal{Z}$. This input density matching was exploited in many previous approaches~\cite{dom_conf,grl16,bousmalis2016domain,tzeng2017adversarial}, and typically accomplished through adversarial learning~\cite{gan14} or the maximum mean discrepancy~\cite{mmd}.

\textbf{Tighter bound.} Recently, \cite{saito2018} exploited the tighter bound (\ref{eq:thm_2010}) under the assumption that $\mathcal{H}$ is restricted to classifiers with small errors on $S$. Consequently, $d_S(h,h')$ becomes negligible as any two $h,h'\in\mathcal{H}$  agree on the source domain. The supremum in (\ref{eq:thm_2010}), interpreted as the \textit{Maximum Classifier Discrepancy (\textbf{MCD})}, reduces to:
\begin{equation}
\sup_{h,h' \in \mathcal{H}} \mathbb{E}_{({\bf x},y)\sim p_T({\bf x},y)} [ \mathbb{I}(h({\bf z}) \neq h'({\bf z})) ]. 
\label{eq:max_discrep}
\end{equation}
Named \textbf{MCDA}, \cite{saito2018} aims to minimize (\ref{eq:max_discrep}) directly via adversarial-cooperative learning of two deep classifier networks $h({\bf z})$ and $h'({\bf z})$. For the source domain data, these two classifiers and ${\bf G}$ aim to minimize the classification errors cooperatively. An adversarial game is played in the target domain: $h$ and $h'$ aim to be maximally discrepant, whereas ${\bf G}$ seeks to minimize the discrepancy\footnote{See the Supplementary Material for further technical details.}.



\section{Our Approach}\label{sec:ours}

\textbf{Overview.} We adopt the \textbf{MCD} principle, but propose a more systematic and effective way to achieve hypothesis consistency, instead of the difficult minimax optimization. Our idea is to adopt a Bayesian framework to induce the hypothesis space. 
Specifically, we build a Gaussian process classifier model~\cite{gpml_book} on top of the share space. The \textbf {GP} posterior inferred from the source domain data naturally defines our hypothesis space $\mathcal{H}$. We then optimize the embedding ${\bf G}$ and the kernel of the \textbf {GP} so that the posterior hypothesis distribution leads to consistent (least discrepant) class predictions most of the time, resulting in reduction of (\ref{eq:max_discrep}). The details are described in the below.

\subsection{GP-endowed Maximum Separation Model}

We consider a multi-class Gaussian process classifier defined on $\mathcal{Z}$: there are 
$K$ underlying latent functions ${\bf f}(\cdot) := \{f_j(\cdot)\}_{j=1}^K$, a priori independently \textbf {GP} distributed, namely 
\begin{equation}
P({\bf f}) = \prod_{j=1}^K P(f_j), \ 
  \ f_j \sim \mathcal{GP}\big( 0, k_j(\cdot,\cdot) \big), 
 \label{eq:gp_prior}
\end{equation}
where each $k_j$ is a covariance function of $f_j$, defined on $\mathcal{Z}\times\mathcal{Z}$. 
For an input point ${\bf z}\in\mathcal{Z}$, we regard $f_j({\bf z})$ as the model's confidence toward class $j$, leading to the class prediction rule:
\vspace{+0.5em}
\begin{equation}
\textrm{class}({\bf z}) = \arg\max_{1\leq j\leq K} f_j({\bf z}).
\label{eq:gp_class_pred}
\end{equation}
We use the softmax likelihood model, 
\begin{equation}
P(y=j|{\bf f}({\bf z})) = \frac{ e^{f_j({\bf z})} } { \sum_{r=1}^K e^{f_r({\bf z})} }, \ \ \textrm{for} \ j=1,\dots,K.
 \label{eq:gp_lik}
\vspace{+0.5em}
\end{equation}

\noindent\textbf{Source-driven $\boldsymbol{\mathcal{H}}$ Prior.} The labeled source data, $\mathcal{D}_S$, induces a posterior distribution on the latent functions ${\bf f}$, 
\begin{equation}
p({\bf f}|\mathcal{D}_S) \propto p({\bf f}) \cdot \prod_{i=1}^{N_S} P(y^S_i|{\bf f}({\bf z}^S_i)),
\label{eq:gp_posterior}
\end{equation}
where ${\bf z}^S_i = {\bf G}({\bf x}^S_i)$. The key idea is to use (\ref{eq:gp_posterior}) to define our hypothesis space $\mathcal{H}$. The posterior places most of its probability mass on those ${\bf f}$ that attain high likelihood scores on $S$ while being smooth due to the \textbf {GP} prior.  
It should be noted that we used the term {\em prior} of the hypothesis space $\mathcal{H}$ that is induced from the {\em posterior} of the latent functions ${\bf f}$. We use the $\mathcal{H}$ prior and the posterior of ${\bf f}$ interchangeably.

Note that due to the non-linear/non-Gaussian likelihood (\ref{eq:gp_lik}), exact posterior inference is intractable, and one has to resort to  approximate inference. We will discuss an approach for efficient variational approximate inference 
in Sec.~\ref{sec:vi}. 
For the exposition here, let us assume that the posterior distribution is accessible. 


\noindent\textbf{Target-driven Maximally Consistent Posterior.} While $\mathcal{D}_S$ serves to induce the prior of $\mathcal{H}$, $\mathcal{D}_T$ will be used to reshape this prior.  According to \textbf{MCD}, we want this hypothesis space to be shaped in the following way: for each target domain point ${\bf z} = {\bf G}({\bf x})$, ${\bf x}\sim T$, the latent function values ${\bf f}({\bf z})$ sampled from the posterior (\ref{eq:gp_posterior}) should lead to the class prediction (made by (\ref{eq:gp_class_pred})) that is as consistent as possible across the samples. 

\begin{figure}[t]
\begin{center}
  \includegraphics[trim = 0mm 0mm 26mm 0mm, clip, scale=0.435]{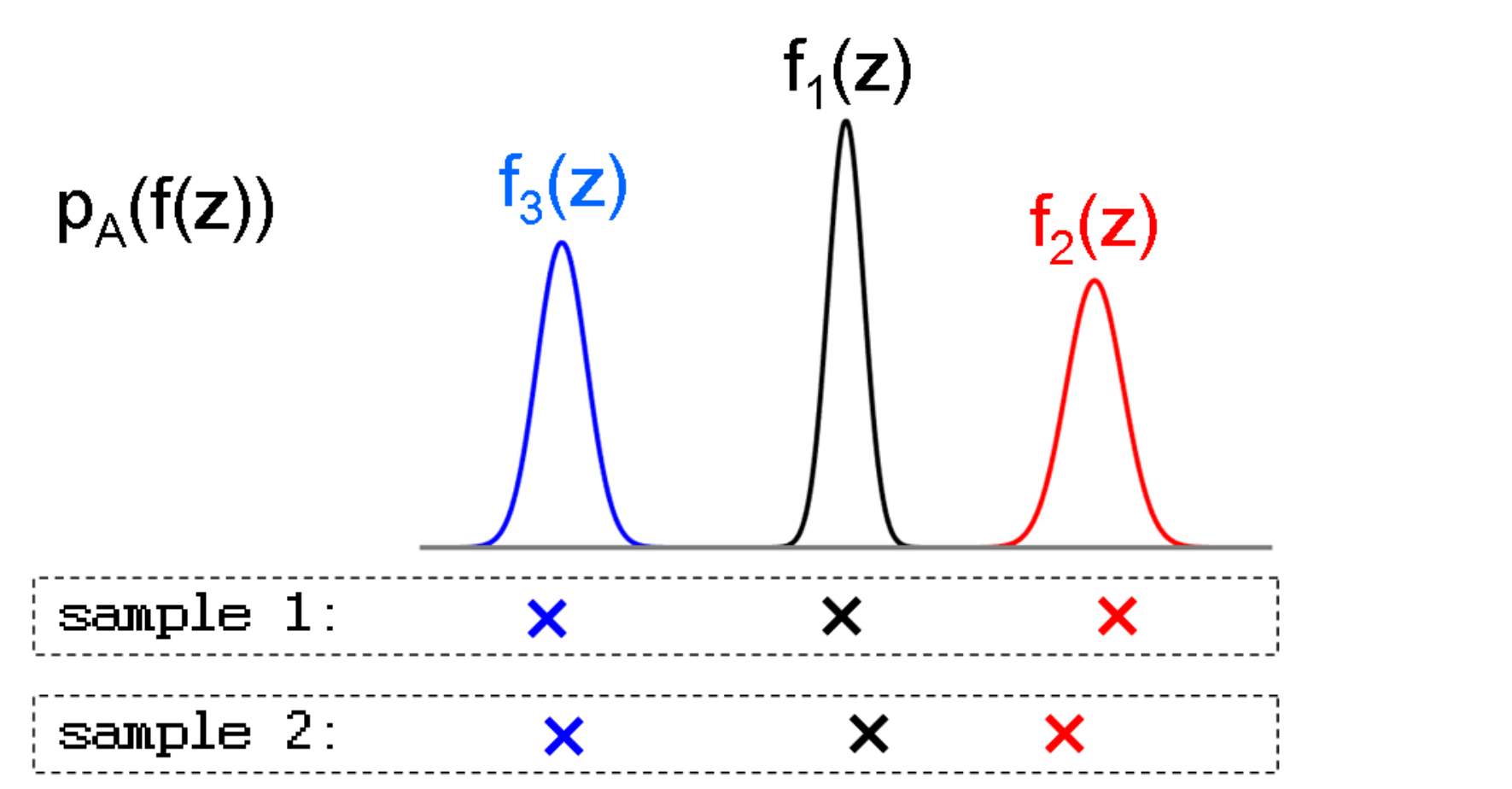} \ 
  \includegraphics[trim = 0mm 0mm 26mm 0mm, clip, scale=0.435]{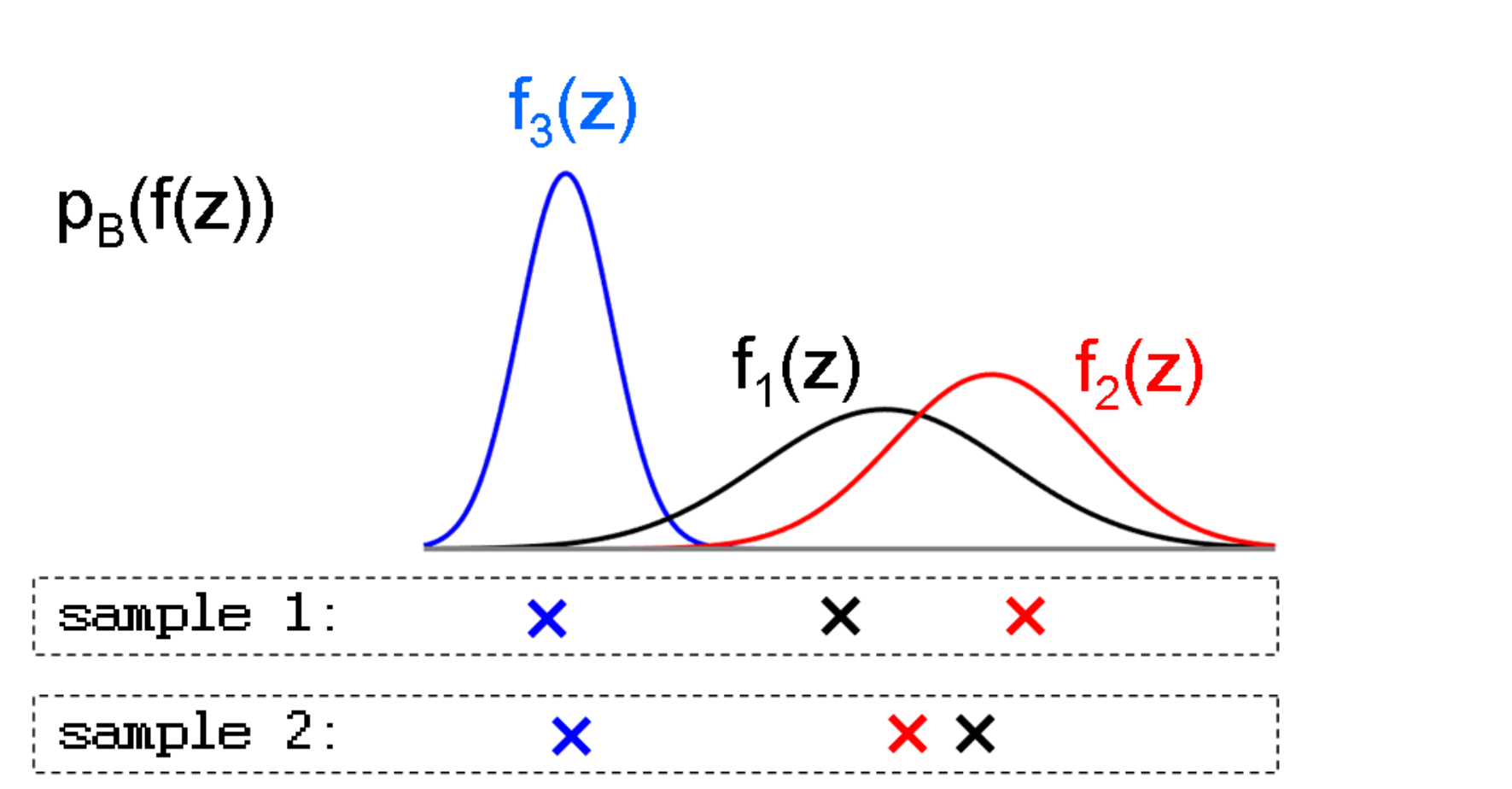}
\caption{Illustration of ideal ($p_A$) and problematic ($p_B$)  posteriors at some fixed point ${\bf z}$ in the target domain. For each posterior, we also depict two plausible samples (marked as crosses). In $p_A$, most samples ${\bf f}({\bf z})$, including the two shown, are consistent in deciding the class label (class $2$, red, predicted in this case). 
On the other hand, in $p_B$ where $f_1({\bf z})$ and $f_2({\bf z})$ have considerable overlap, there is significant chance of different predictions: class $2$ for the first sample and class $1$ for the second.}
\label{fig:illustrate}
\end{center}
\vspace{-1.2em}
\end{figure}

This is illustrated in Fig.~\ref{fig:illustrate}. Consider two different $\mathcal{H}$ priors $p_A$ and $p_B$ at a point ${\bf z}$,  $p_A({\bf f}({\bf z}))$ and $p_B({\bf f}({\bf z}))$, where for brevity we drop the conditioning on $\mathcal{D}_S$ in notation. The class cardinality is $K=3$. For simplicity, we assume that the latent functions $f_j$'s are independent from each other. 
Fig.~\ref{fig:illustrate} shows that the distributions of $f_j$'s are well-separated from each other in $p_A$, yet overlap significantly in $p_B$. Hence, there is a strong chance for the class predictions to be inconsistent in $p_B$ (identical ordering of colored samples below figure), but consistent in $p_A$. This means that the hypothesis space induced from $p_B$ contains highly discrepant classifiers, whereas most classifiers in the hypothesis space of $p_A$ agree with each other (least discrepant). In other words, the maximum discrepancy principle translates into the {\em maximum posterior separation} in our Bayesian \textbf {GP} framework. 

We describe how this goal can be properly formulated. First we consider the posterior of ${\bf f}$ to be approximated as an independent Gaussian\footnote{This choice conforms to the variational density family we choose in Sec.~\ref{sec:vi}.}. For any target domain point ${\bf z}\sim T$ and each $j=1,\dots,K$ let the mean and the variance of the $\mathcal{H}$ prior in (\ref{eq:gp_posterior}) be: 
\begin{eqnarray}
\mu_j({\bf z}) &:=& \int f_j({\bf z}) \ 
  p\big( f_j({\bf z})|\mathcal{D}_S, {\bf z} \big) \ d f_j({\bf z}), \label{eq:gp_post_mean} \\
\sigma_j^2({\bf z}) &:=& \int ( f_j({\bf z}) - \mu_j({\bf z}) )^2 \  
  p\big( f_j({\bf z})|\mathcal{D}_S, {\bf z} \big) \ d f_j({\bf z}). \label{eq:gp_post_stdev} \raisetag{1.2\baselineskip}
\vspace{+0.5em}
\end{eqnarray}

The maximum-a-posterior (MAP) class prediction by the model is denoted by $j^* = \arg\max_{1 \leq j \leq K} \mu_j({\bf z})$. As we seek to avoid fluctuations in class prediction $j^*$ across samples, we consider the worst scenario where even an  unlikely (e.g., at $5\%$ chance level) sample from $f_j({\bf z})$, $j$ other than $j^*$, cannot overtake $\mu_{j^*}({\bf z})$. That is, we seek
\vspace{+0.5em}
\begin{equation}
    \mu_{j^*}({\bf z}) - \alpha \sigma_{j^*}({\bf z}) 
    \geq \max_{j\neq j^*} \big( \mu_j({\bf z}) + \alpha \sigma_j({\bf z}) \big),
\label{eq:post_sep_ineq}
\vspace{+0.5em}
\end{equation}
where $\alpha$ is the normal cutting point for the least chance (e.g., $\alpha=1.96$ if $2.5\%$ one-side is considered). 

While this should hold for most samples, it will not hold for all. We therefore introduce an additional slack $\xi\geq 0$ to relax the desideratum. Furthermore, for ease of  optimization\footnote{We used the \texttt{topk()} function in \texttt{PyTorch} to compute the largest and the second largest elements. 
The function allows automatic gradients.}, we impose slightly stricter constraint than (\ref{eq:post_sep_ineq}), leading to the final constraint:
\begin{equation}
\max_{1\leq j \leq K} \mu_j({\bf z}) \geq 1 + \max_{j \neq j^*} \mu_j({\bf z}) + \alpha \max_{1\leq j \leq K} \sigma_j({\bf z}) - \xi({\bf z}).
\label{eq:max_marg_ineq}
\end{equation}
A constant, $1$ here, was added to normalize the scale of $f_j$'s.

Our objective now is to find such embedding ${\bf G}$, \textbf {GP} kernel parameters $k$, and minimal slacks $\xi$, to impose (\ref{eq:max_marg_ineq}). Equivalently, we pose it as the following optimization problem, for each ${\bf z}\sim T$:
\begin{equation} \small
\min_{{\bf G},k} \bigg( \max_{j \neq j^*} \mu_j({\bf z}) - \max_{1\leq j \leq K} \mu_j({\bf z}) +
    1 + \alpha \max_{1\leq j \leq K} \sigma_j({\bf z}) \bigg)_+
\label{eq:max_marg_obj}
\end{equation}
with $(a)_+ = \max(0,a)$.

Note that (\ref{eq:max_marg_ineq}) and (\ref{eq:max_marg_obj}) are reminiscent of the large-margin classifier learning in traditional supervised learning~\cite{vapnik_book98}. In contrast, we replace the ground-truth labels with the {\em the most confidently} predicted labels by our model since the target domain is unlabeled. This aims to place class boundaries in low-density regions, in line with {\em entropy minimization} or {\em max-margin confident prediction} principle of classical semi-supervised learning~\cite{ssem04,semisup_book}.

In what follows, we describe an approximate, scalable \textbf {GP} posterior inference, where we combine the variational inference optimization with the aforementioned posterior maximum separation criterion (\ref{eq:max_marg_obj}).

\subsection{Variational Inference with Deep Kernels}\label{sec:vi}

We describe our scalable variational inference approach to approximate the posterior (\ref{eq:gp_posterior}). Although there are scalable \textbf {GP} approximation schemes based on the random feature expansion~\cite{rf_fourier} and the pseudo/induced inputs~\cite{quinonero05,snelson06,titsias09,dezfouli15}, here we adopt the {\em deep kernel} trick~\cite{deep_kernel,dkl16} to exploit the deeply structured features. The main idea is to model an explicit finite-dimensional feature space mapping to define a covariance function. Specifically, we consider a nonlinear feature mapping ${\boldsymbol\phi}:\mathcal{Z} \to \mathbb{R}^d$ such that the covariance function is defined as an inner product in a feature space, namely $k({\bf z},{\bf z}') := {\boldsymbol\phi}({\bf z})^\top {\boldsymbol\phi}({\bf z}')$, where we model ${\boldsymbol\phi}(\cdot)$ as a deep neural network.
A critical advantage of explicit feature representation is that we turn the non-parametric \textbf {GP} into a parametric Bayesian model. 
As a consequence, all inference operations in the non-parametric \textbf {GP} reduce to computationally more efficient parametric ones, avoiding the need to store the Gram matrix of the entire training data set, as well as its inversion. 

Formally, we consider $K$ latent functions modeled as $f_j({\bf z}) = {\bf w}_j^\top {\boldsymbol \phi}({\bf z})$ with ${\bf w}_j \sim \mathcal{N}({\bf 0},{\bf I})$ independently for $j=1,\dots,K$. We let ${\bf W} = [{\bf w}_1,\dots,{\bf w}_K]^\top$. Note that the feature function ${\boldsymbol \phi}(\cdot)$ is shared across classes to reduce the number of parameters and avoid overfitting.
The parameters of the deep model that represents ${\boldsymbol \phi}(\cdot)$ serve as \textbf {GP} kernel parameters, since $\textrm{Cov}(f({\bf z}),f({\bf z}')) = \textrm{Cov}({\bf w}^\top {\boldsymbol \phi}({\bf z}), {\bf w}^\top  {\boldsymbol \phi}({\bf z}')) =  {\boldsymbol \phi}({\bf z})^\top {\boldsymbol \phi}({\bf z}') = k({\bf z},{\bf z}')$. 
Consequently, the source-driven $\mathcal{H}$ prior (\ref{eq:gp_posterior}) becomes
\begin{equation}
p({\bf W}|\mathcal{D}_S) \propto \prod_{j=1}^K \mathcal{N}({\bf w}_j; {\bf 0}, {\bf I}) \cdot \prod_{i=1}^{N_S} P(y^S_i|{\bf W} {\boldsymbol \phi}({\bf z}^S_i)). 
\label{eq:w_posterior}
\end{equation}

Since computing (\ref{eq:w_posterior}) is intractable, we introduce a variational density $q({\bf W})$ to approximate it. We assume a fully factorized Gaussian, 
\begin{equation}
q({\bf W}) = \prod_{j=1}^K \mathcal{N}({\bf w}_j; {\bf m}_j, {\bf S}_j),
\label{eq:inf_q}
\end{equation}
where ${\bf m}_j\in\mathbb{R}^d$ and ${\bf S}_j\in\mathbb{R}^{d\times d}$ constitute the variational parameters. We further let ${\bf S}_j$'s be diagonal matrices. To have $q({\bf W}) \approx p({\bf W}|\mathcal{D}_S)$, we use the following fact that the marginal log-likelihood can be lower bounded: 
\vspace{+0.5em}
\begin{equation}
\log P \Big( \{y^S_i\}_{i=1}^{N_S} \ \Big| \ \{{\bf z}^S_i\}_{i=1}^{N_S}, 
    {\boldsymbol \phi}(\cdot) \Big) 
  \geq \textrm{ELBO},
\label{eq:ineq_elbo}
\vspace{+0.5em}
\end{equation}
where the evidence lower-bound (ELBO) is defined as:
\vspace{+0.5em}
\begin{equation}
\textrm{ELBO} :=
  \sum_{i=1}^{N_S} \mathbb{E}_{q({\bf W})} \big[ 
    \log P(y^S_i|{\bf W} {\boldsymbol \phi}({\bf z}^S_i)) \big]  - 
  \sum_{j=1}^K \textrm{KL} \big( q({\bf w}_j) \ || \ \mathcal{N}({\bf w}_j; {\bf 0}, {\bf I}) \big), 
\label{eq:elbo}
\vspace{+0.5em}
\end{equation}
with the likelihood stemming from (\ref{eq:gp_lik}).
As the gap in (\ref{eq:ineq_elbo}) is the KL divergence between $q({\bf W})$ and the true posterior $p({\bf W}|\mathcal{D}_S)$, increasing the $\textrm{ELBO}$ wrt the variational parameters $\{({\bf m}_j, {\bf S}_j)\}$ brings $q({\bf W})$ closer to the true posterior. 
Raising the $\textrm{ELBO}$ wrt the \textbf{GP} kernel parameters (i.e., the parameters of ${\boldsymbol \phi}$) 
and the embedding\footnote{Note that the inputs ${\bf z}$ also depend on ${\bf G}$.} ${\bf G}$ can potentially improve the marginal likelihood (i.e., the left hand side in (\ref{eq:ineq_elbo})). 

In optimizing the $\textrm{ELBO}$ (\ref{eq:elbo}), the KL term (denoted by KL) can be analytically derived as 
\begin{equation}
\textrm{KL} = \frac{1}{2} \sum_{j=1}^K \big( \ 
  \textrm{Tr}({\bf S}_j) + ||{\bf m}_j||_2^2 - \log\det({\bf S}_j) - d \ \big).
\label{eq:kl} 
\end{equation}
However, there are two key challenges: the log-likelihood expectation over $q({\bf W})$ does not admit a closed form, and one has to deal with large $N_S$. To address the former, we adopt Monte-Carlo estimation using $M$ iid samples $\{{\bf W}^{(m)}\}_{m=1}^M$ from $q({\bf W})$, where the samples are expressed in terms of the variational parameters (i.e., the reparametrization trick~\cite{vae14}) to facilitate optimization. That is, for each $j$ and $m$, 
\vspace{+0.5em}
\begin{equation}
{\bf w}_j^{(m)} = {\bf m}_j + {\bf S}_j^{1/2}  {\boldsymbol \epsilon}^{(m)}_{j}, 
    \ \ \ \ \ {\boldsymbol \epsilon}^{(m)}_{j} \sim \mathcal{N}({\bf 0},{\bf I}).
\label{eq:reparam_W}
\vspace{+0.5em}
\end{equation}
%
For the latter issue, we use stochastic optimization with a random mini-batch $B_S \subset \mathcal{D}_S$. That is, we optimize the sample estimate of the log-likelihood defined as:
\begin{equation}
\textrm{LL} = \frac{1}{M} \sum_{m=1}^M \frac{N_S}{|B_S|} \sum_{i \in B_S}
    \log P(y^S_i | {\bf W}^{(m)} {\boldsymbol \phi}({\bf z}^S_i)).
\label{eq:ll} 
\end{equation}

\subsection{Optimization Strategy}\label{sec:final_optim}

Now we combine the maximum posterior separation criterion in (\ref{eq:max_marg_obj}) with the variational inference discussed in the previous section to arrive at the comprehensive optimization task.

Our approximate posterior (\ref{eq:inf_q}) leads to closed-form expressions for $\mu_j({\bf z})$ and $\sigma_j({\bf z})$ in (\ref{eq:gp_post_mean}--\ref{eq:gp_post_stdev}) as follows: 
\vspace{+0.5em}
\begin{equation}
\mu_j({\bf z}) \approx {\bf m}_j^\top {\boldsymbol \phi}({\bf z}), \ \ \ \  
\sigma_j({\bf z}) \approx 
  \big( {\boldsymbol \phi}({\bf z})^\top {\bf S}_j {\boldsymbol \phi}({\bf z}) \big)^{1/2}. 
  \label{eq:approx_post_mean_stdev}
\vspace{+0.5em}
\end{equation}
With $q({\bf W})$ fixed, we rewrite our posterior maximum separation loss in (\ref{eq:max_marg_obj}) as follows.  We consider stochastic optimization with a random mini-batch $B_T\subset\mathcal{D}_T = \{{\bf z}_i^T\}_{i=1}^{N_T}$ sampled from the target domain data.
\begin{equation}
\textrm{MS} := \frac{1}{|B_T|} \sum_{i\in B_T} \bigg( 
  \max_{j \neq j^*} {\bf m}_j^\top {\boldsymbol \phi}({\bf z}^T_i) - 
  \max_{1\leq j \leq K} {\bf m}_j^\top {\boldsymbol \phi}({\bf z}^T_i) 
\ + \ 1 \ + \ \alpha \max_{1\leq j \leq K} 
    \big( {\boldsymbol \phi}({\bf z}^T_i)^\top {\bf S}_j {\boldsymbol \phi}({\bf z}^T_i) \big)^{1/2} 
\bigg)_+ 
\label{eq:ms}
\end{equation}

Combining all objectives thus far, our algorithm\footnote{
In the algorithmic point of view, our algorithm can be viewed as a {\em max-margin Gaussian process classifier} on the original input space $\mathcal{X}$ without explicitly considering a shared space $\mathcal{Z}$. 
For further details about this connection, the reader is encouraged to refer to the Supplementary Material. 
} can be summarized as the following two optimizations alternating with each other: 
\begin{mdframed}
\begin{itemize}
\item $\min_{\{{\bf m}_j,{\bf S}_j\}} \  -\textrm{LL} + \textrm{KL}$ \ \ \ \ \ 
(variational inference)
\item $\min_{{\bf G},k} 
\  -\textrm{LL} + \textrm{KL} + \lambda \cdot \textrm{MS}$ \ \ \ \ (model selection)
\end{itemize}
\end{mdframed}
where $\lambda$ is the impact of the maximum separation loss (e.g., $\lambda=10.0$).
%


\section{Related Work}\label{sec:related}

There has been extensive prior work on domain adaptation~\cite{csurka2017comprehensive}. 
Recent  approaches  have  focused  on transferring  deep  neural  network  representations  from  a labeled source dataset to an unlabeled target domain by matching the distributions of features between different domains, aiming to extract domain-invariant features~\cite{rebuffi2017learning,benaim2017one,courty2017joint,motiian2017few,saito2017asymmetric,Zhang_2017_CVPR,Yan_2017_CVPR,bousmalis2017unsupervised,mancini2018boosting,rozantsev2018residual}.
To this end, it is critical to first define a measure of distance (divergence) between source and target distributions. One popular measure is the non-parametric Maximum Mean Discrepancy (\textbf{MMD}) (adopted by~\cite{bousmalis2017unsupervised,zellinger2017central,long2014transfer}), which measures the distance between the sample means of the two domains in the reproducing Kernel Hilbert Space (\textbf{RKHS}) induced by a pre-specified kernel. The deep Correlation Alignment (\textbf{CORAL}) method~\cite{sun2016deep} attempted to match the sample mean and covariance of the source/target distributions, while it was further generalized to potentially infinite-dimensional feature spaces in~\cite{zhang2018aligning} to effectively align the \textbf{RKHS} covariance matrices (descriptors) across domains. 

The Deep Adaptation Network (\textbf{DAN})~\cite{long2015learning} applied \textbf{MMD} to layers embedded in a \textbf{RKHS} to match higher order moments of the two distributions more effectively. The Deep Transfer Network (\textbf{DTN})~\cite{zhang2015deep} achieved alignment of source and target distributions using two types of network layers based on the \textbf{MMD} distance: the shared feature extraction layer that can learn a subspace that matches the marginal distributions of the source and the target samples, and the discrimination layer that can match the conditional distributions by classifier transduction.

Many recent \textbf{UDA} approaches leverage deep neural networks with the adversarial training strategy~\cite{rebuffi2017learning,benaim2017one,courty2017joint,motiian2017few,saito2017asymmetric,Zhang_2017_CVPR}, which allows the learning of feature representations to be simultaneously discriminative for the labeled source domain data and indistinguishable between source and target domains. For instance, \cite{ganin2014unsupervised} proposed a technique called the Domain-Adversarial Training of Neural Networks (\textbf{DANN}), which allows the network to learn domain invariant representations in an adversarial fashion by adding an auxiliary domain classifier and back-propagating inverse gradients. The Adversarial Discriminative Domain  Adaptation (\textbf{ADDA})~\cite{tzeng2017adversarial} first learns a discriminative feature subspace using the labeled source samples. Then, it encodes the target data to this subspace using an asymmetric transformation learned through a domain-adversarial loss. The \textbf{DupGAN}~\cite{hu2018duplex} proposed a \textbf{GAN}-like model~\cite{gan14} with duplex discriminators to restrict the latent representation to be domain invariant while its category information being preserved.

In parallel, within the shared-latent space framework, \cite{liu2017unsupervised} proposed an unsupervised image-to-image translation (\textbf{UNIT}) framework based on the \textbf{Coupled GAN}s~\cite{NIPS2016_6544}.  Another interesting idea is the pixel-level domain adaptation method (\textbf{PixelDA})~\cite{bousmalis2017unsupervised} where they imposed alignment of distributions not in the feature space but directly in the raw pixel space via the adversarial approaches. The intention is to adapt the source samples as if they were drawn from the target domain, while maintaining the original content. Similarly, \cite{murez2017image} utilized the \textbf{CycleGAN}~\cite{zhu2017unpaired} to constrain the features extracted by the encoder network to reconstruct the images in both domains. In \cite{sankaranarayanan2017generate}, they proposed a joint adversarial discriminative approach that can transfer the information of the target distribution to the learned embedding using a generator-discriminator pair.

\section{Experimental Results}\label{sec:expmt}
We compare the proposed method with state-of-the-art on standard benchmark datasets.  Digit classification task consists of three datasets, containing ten digit classes: \textbf{MNIST}~\cite{lecun1998gradient}, \textbf{SVHN}~\cite{netzer2011reading}, \textbf{USPS}~\cite{tzeng2017adversarial}. 
We also evaluated our method on the traffic sign datasets, Synthetic Traffic Signs (SYN SIGNS)~\cite{moiseev2013evaluation} and the German Traffic Signs Recognition Benchmark~\cite{stallkamp2011german} (GTSRB), which contain 43 types of signs. Finally, we report performance on  \textbf{VisDA} object classification dataset~\cite{peng2017visda} with more than 280K images across twelve categories ( the details of the datasets are available in the Supplementary Material).  \autoref{amazon} illustrates image samples from different datasets and domains.

\begin{figure}[t]
 \centering
 \begin{subfigure}[t]{0.3\linewidth}
    \includegraphics[trim = -70mm 2mm 5mm 0mm, clip, scale=0.112]{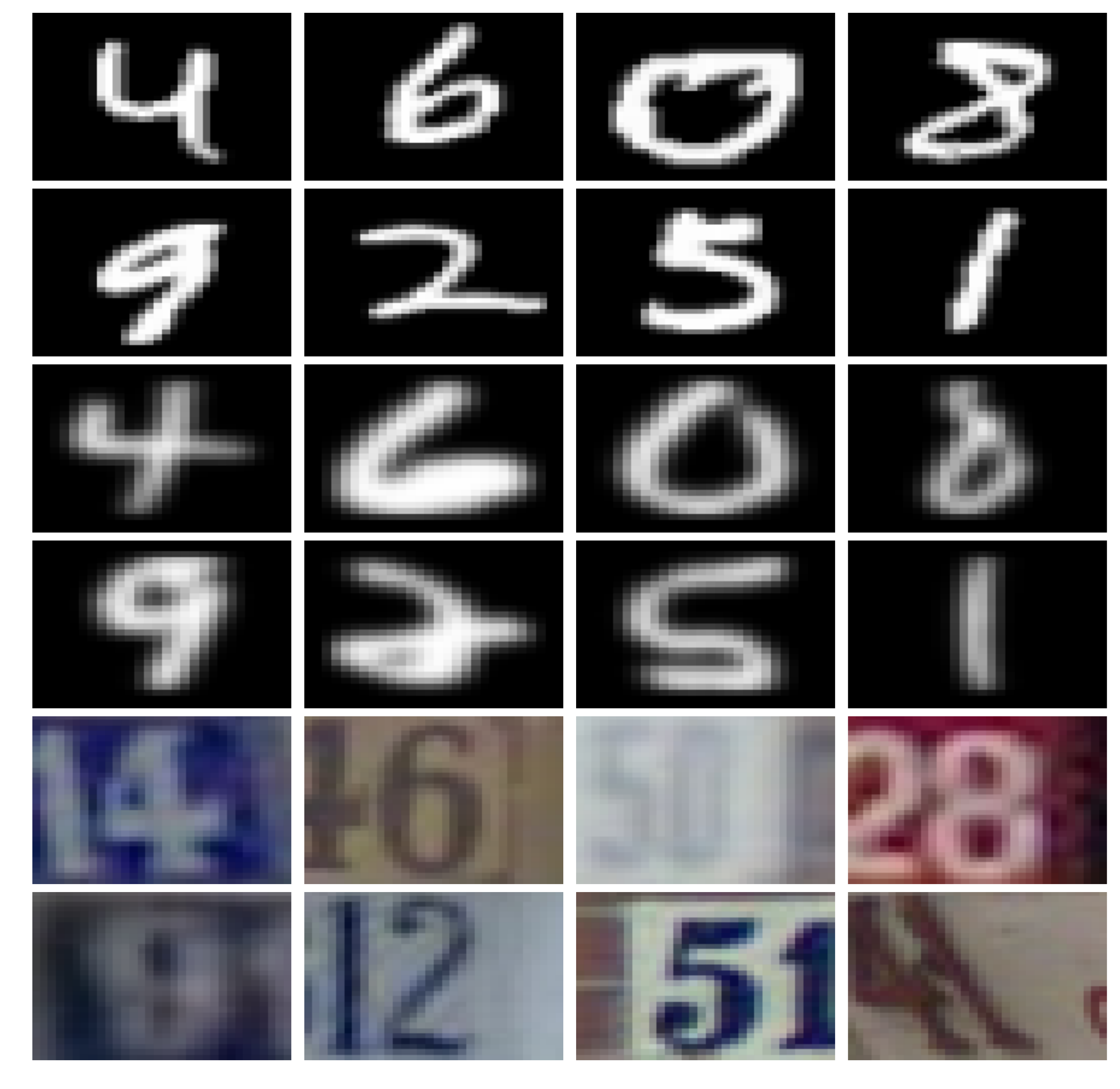} 
    \caption{Digits.}
 \end{subfigure} \
 \begin{subfigure}[t]{0.3\linewidth}
    \includegraphics[trim = -70mm 2mm 5mm 0mm, clip, scale=0.112]{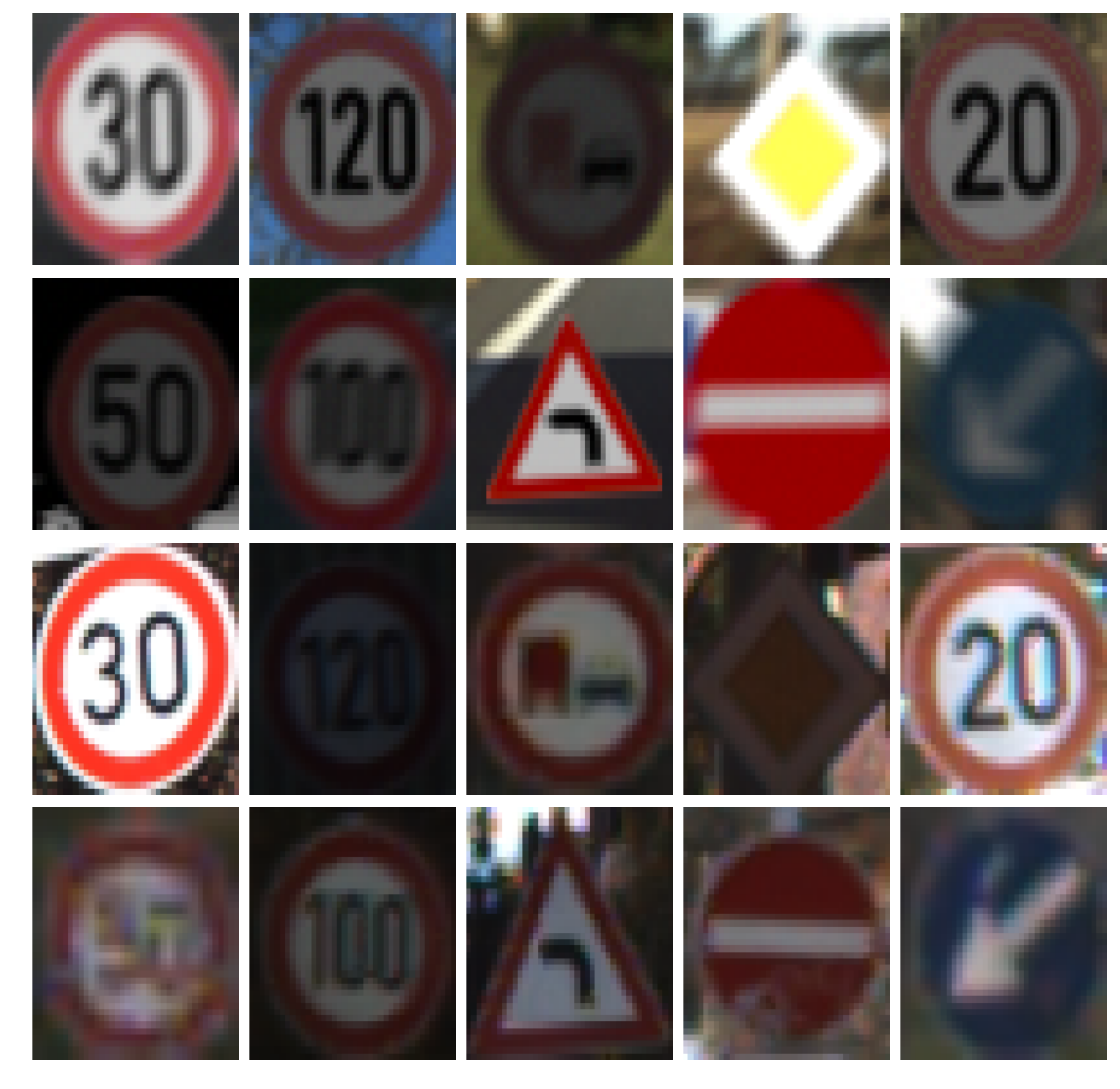}
    \caption{Traffic Signs.}
  \end{subfigure} \
  \begin{subfigure}[t]{0.3\linewidth}
    \includegraphics[trim = -60mm 2mm 5mm 0mm, clip, scale=0.112]{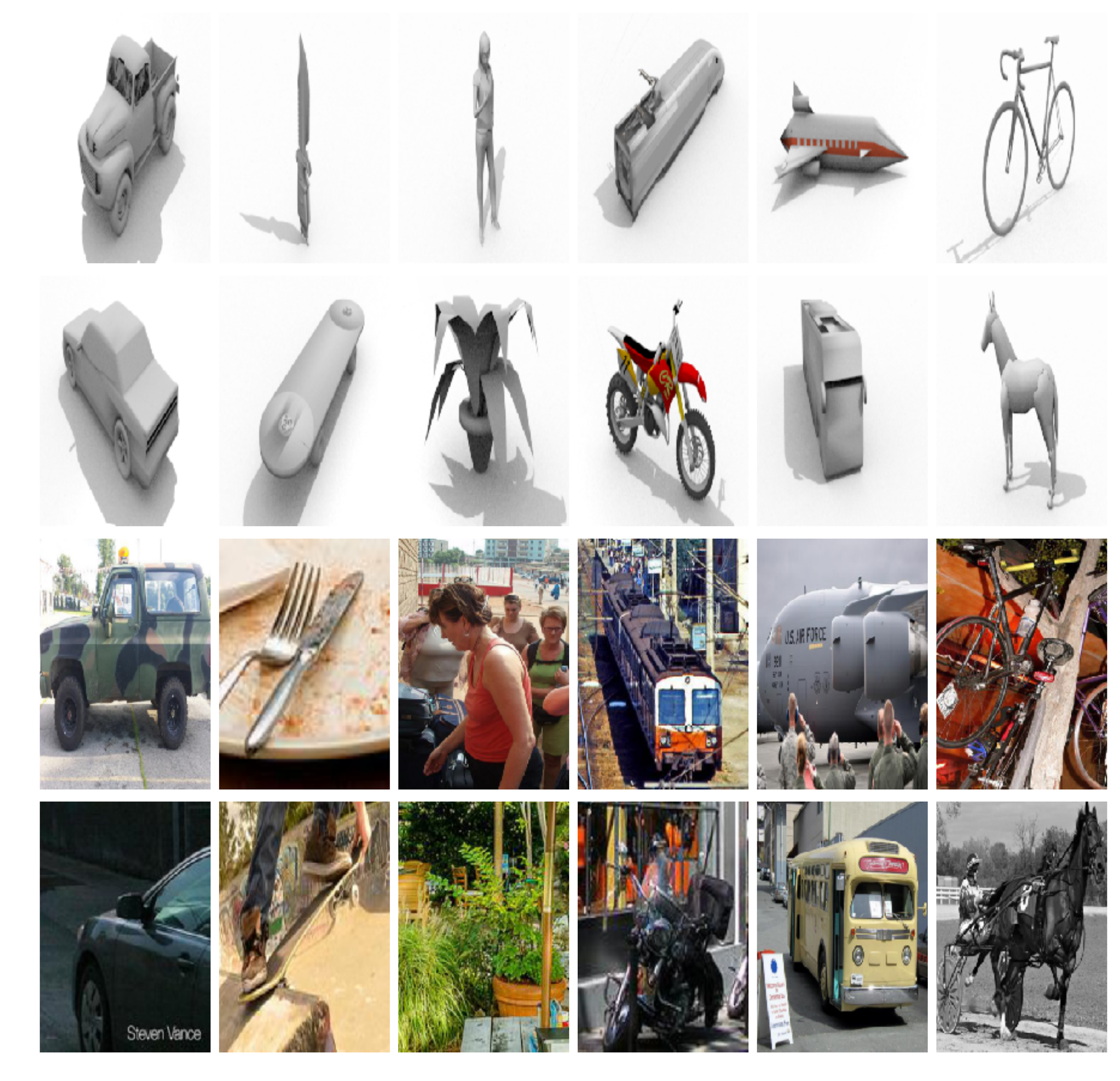}
    \caption{VisDA.}
   \end{subfigure}
\vspace{-0.8em}
\caption{Example images from benchmark datasets. (a) Samples from MNIST, USPS, and SVHN datasets. (b) Samples from SYN SIGNS (first two rows), and GTSRB.
}
\label{amazon}
\vspace{-0.8em}
\end{figure}

We evaluate the performance of all methods with the classification accuracy score. We used ADAM~\cite{kingma2014adam} for training; the learning rate was set to $0.0002$ and momentum to $0.5$ and $0.999$. We used batches of size $32$ from each domain, and the input images were mean-centered. The hyper-parameters are empirically set as $\lambda =50.0, \alpha = 2.0$. The sensitivity w.r.t. hyperparameters $\lambda$ and $\alpha$ will
be discussed in \autoref{ablation}. We also used the same network structure as~\cite{saito2018}. Specifically, we employed the CNN architecture used in~\cite{ganin2014unsupervised} and~\cite{bousmalis2017unsupervised} for digit and traffic sign datasets and used ResNet101~\cite{he2016deep} model pre-trained on Imagenet~\cite{deng2009imagenet}. We added batch normalization to each layer in these models. Quantitative  evaluation  involves a comparison
of the performance of our model to previous works and to
“\textbf{Source Only}” that do not use any  domain  adaptation. For "\textbf{Source Only}" baseline, we  train  models on  the  unaltered  source  training  data  and  evaluate on the target test data. The training details
for comparing methods are available in our Supplementary
material due to the space limit.

\subsection{Results on Digit and Traffic Signs datasets}
We show the accuracy of different methods in \autoref{tb:exp_digit}. It can be seen the proposed method outperformed competitors in all settings confirming consistently better generalization of our model over target data. This is partially due to combining DNNs and GPs/Bayesian approach. GPs exploit local generalization by locally interpolating between neighbors~\cite{bengio2013representation}, adjusting the target functions rapidly in the presence of training data.  DNNs have good generalization capability for unseen input configurations by learning multiple levels of distributed representations. The results demonstrate \textbf{GPDA} can improve generalization performance by adopting both of these advantages.

\subsection{Results on VisDA dataset}
Results for this experiment are summarized in \autoref{tb:visda}. We observe that our \textbf{GPDA} achieved, on average, the best performance compared to other competing methods. Due to vastly varying difficulty of classifying different categories of objects, in addition to reporting the average classification accuracy we also report the average rank of each method over all objects (the lower rank, the better). The higher performance of \textbf{GPDA} compared to other methods is mainly attributed to modeling the classifier as a random function and consequently incorporating the classifier uncertainty (variance of the prediction) into the proposed loss function, \autoref{eq:ms}. The image structure for this dataset is more complex than that of digits, yet our method exhibits very strong performance even under such challenging conditions.\\
Another key observation is that some of the competing methods (e.g., \textbf{MMD}, \textbf{DANN}) perform worse than the source-only model in classes such as car and plant, while \textbf{GPDA} and \textbf{MCDA} performed better across all classes, which clearly demonstrates the effectiveness of the \textbf{MCD} principle.

\begin{table}[t]
\begin{center}
\scalebox{0.9}{
  \begin{tabular}{|l|c|c|c|c|c|c|}
& {\bf SVHN}&{\bf SYNSIG}&{\bf MNIST}&{\bf MNIST$^*$}&{\bf USPS}\\
{\bf METHOD}&to&to&to&to&to\\
 & {\bf MNIST} &{\bf GTSRB}&{\bf USPS}&{\bf USPS$^*$}&{\bf MNIST}\\ \hline
Source Only &67.1&85.1&76.7&79.4&63.4\\\hline
 MMD $\dagger$~\cite{long2015learning}&71.1&91.1&-&81.1&-\\
 DANN $\dagger$~\cite{ganin2014unsupervised}&71.1&88.7&77.1$\pm$1.8&85.1&73.0$\pm$0.2\\
 DSN $\dagger$~\cite{bousmalis2016domain}&82.7&93.1&91.3&-&-\\
  ADDA~\cite{tzeng2017adversarial}&76.0$\pm$1.8&-&89.4$\pm$0.2&-&90.1$\pm$0.8\\
  CoGAN~\cite{NIPS2016_6544}&-&-&91.2$\pm$0.8&-&89.1$\pm$0.8\\
  PixelDA~\cite{bousmalis2017unsupervised}&-&-&-&95.9&-\\
ATDA $\dagger$~\cite{saito2017asymmetric}&86.2&\color{red}{96.1}&-&-&-\\
 ASSC~\cite{haeusser2017associative} &95.7$\pm$1.5&82.8$\pm$1.3&-&-&-\\
 DRCN~\cite{ghifary2016deep}&82.0$\pm$0.1&-&91.8$\pm$0.09&-&73.7$\pm$0.04\\\hline
 MCDA ($n=2$)&94.2$\pm$2.6&93.5$\pm$0.4&92.1$\pm$0.8&93.1$\pm$1.9&90.0$\pm$1.4\\
 MCDA ($n=3$)& 95.9$\pm$0.5&94.0$\pm$0.4&93.8$\pm$0.8&95.6$\pm$0.9&91.8$\pm$0.9\\
 MCDA ($n=4$)&{\color{red}{96.2}}$\pm$0.4&{ 94.4}$\pm$0.3&{\color{red}{94.2}}$\pm$0.7&{\color{red}{96.5}}$\pm$0.3&{\color{red}{94.1}}$\pm$0.3\\
 \hline\hline
 GPDA &{\bf\color{red}{98.2}}$\pm$0.1&{\bf \color{red}{96.19}}$\pm$0.2&{\bf\color{red}{96.45}}$\pm$0.15&{\bf\color{red}{98.11}}$\pm$0.1&{\bf\color{red}{96.37}}$\pm$0.1\\
\end{tabular}}
\vspace{-1.4em}
\end{center}
 \caption{Classification results on the digits and traffic signs datasets (best viewed in color). The
		best score is in bold red, second best in light red. Results are cited from each study. The score of MMD is cited from DSN~\cite{bousmalis2016domain}. $\dagger$ indicates the method used a few labeled target samples as validation, different from our GPDA setting. We repeated each experiment five times and report the average and the standard deviation of the accuracy. The accuracy for MCDA was obtained from classifier $F_1$. $n$ is MCDA's hyper-parameter, which denotes the number of times the feature generator is updated to mimic classifiers.  MNIST$^*$ and USPS$^*$ denote all the training samples were used to train the models.}
 \label{tb:exp_digit}
 \end{table}

\begin{table*}[t]
\begin{center}
\label{my-label}
\scalebox{0.85}{
\begin{tabular}{l||cccccccccccc|c|c}
Method     & plane & bcycl & bus  & car  & horse & knife & mcycl & person & plant & sktbrd & train & truck & mean & Ave. ranking \\\hline
Source Only & 55.1      & 53.3    & 61.9 & 59.1 & 80.6  & 17.9  & 79.7        & 31.2   & 81.0    & 26.5       & 73.5  & 8.5   & 52.4 & 6.67\\\hline
MMD~\cite{long2015learning}      & \color{red}{87.1}      & 63.0      & 76.5 & 42.0   & \color{red}{90.3}  & 42.9  & {\bf \color{red}{85.9}}        & 53.1   & 49.7  & 36.3       & {\bf \color{red}{85.8}}  & 20.7  & 61.1 & 3.84\\
DANN~\cite{ganin2014unsupervised}       & 81.9      & {\bf \color{red}{77.7}}    & 82.8 & 44.3 & 81.2  & 29.5  & 65.1        & 28.6   & 51.9  & { \color{red}{54.6}}       & 82.8  & 7.8   & 57.4& 4.40\\\hline
MCDA ($n=2$)&81.1&55.3&\color{red}{83.6}&{ \color{red}{65.7}}&87.6&72.7&83.1&\color{red}{73.9}&85.3&47.7&73.2&27.1&69.7 & 3.75\\
MCDA ($n=3$)&{\bf \color{red}{90.3}}&49.3&82.1&62.9&{\bf \color{red}{91.8}}&69.4&83.8&72.8&79.8&53.3&81.5&{ \color{red}{29.7}}&70.6 & 3.25\\
MCDA ($n=4$)& 87.0&60.9&{\bf \color{red}{83.7}}&64.0&88.9&{\bf \color{red}{79.6}}&\color{red}{84.7}&{\bf\color{red}{76.9}}&{ \color{red}{88.6}}&40.3&\color{red}{83.0}&25.8&{ \color{red}{71.9}} & \color{red}{2.84}\\
\hline\hline
GPDA (ours)& 83.0&{\color{red}{74.3}}&80.4&{\bf\color{red}{66.0}}&87.6&\color{red}{75.3}&83.8&73.1&{\bf\color{red}{90.1}}&{\bf\color{red}{57.3}}&80.2 &{\bf\color{red}{37.9}}&{\bf\color{red}{73.31}} & {\bf\color{red}{2.50}} 
\end{tabular}}
\end{center}
  \vspace{-1.5em}
\caption{Accuracy of ResNet101 model fine-tuned on the VisDA dataset. Last column shows the average rank of each method over all classes. The best (in bold red), the second best (in red).}
\label{tb:visda}
\end{table*}

\subsection{Ablation Studies}\label{ablation}
\begin{figure}[t]
    \centering
        \begin{subfigure}[t]{.45\linewidth}
            \includegraphics[trim={-2.5cm -0.5cm 0cm 0.8cm}, scale=0.30,valign=t
            ]{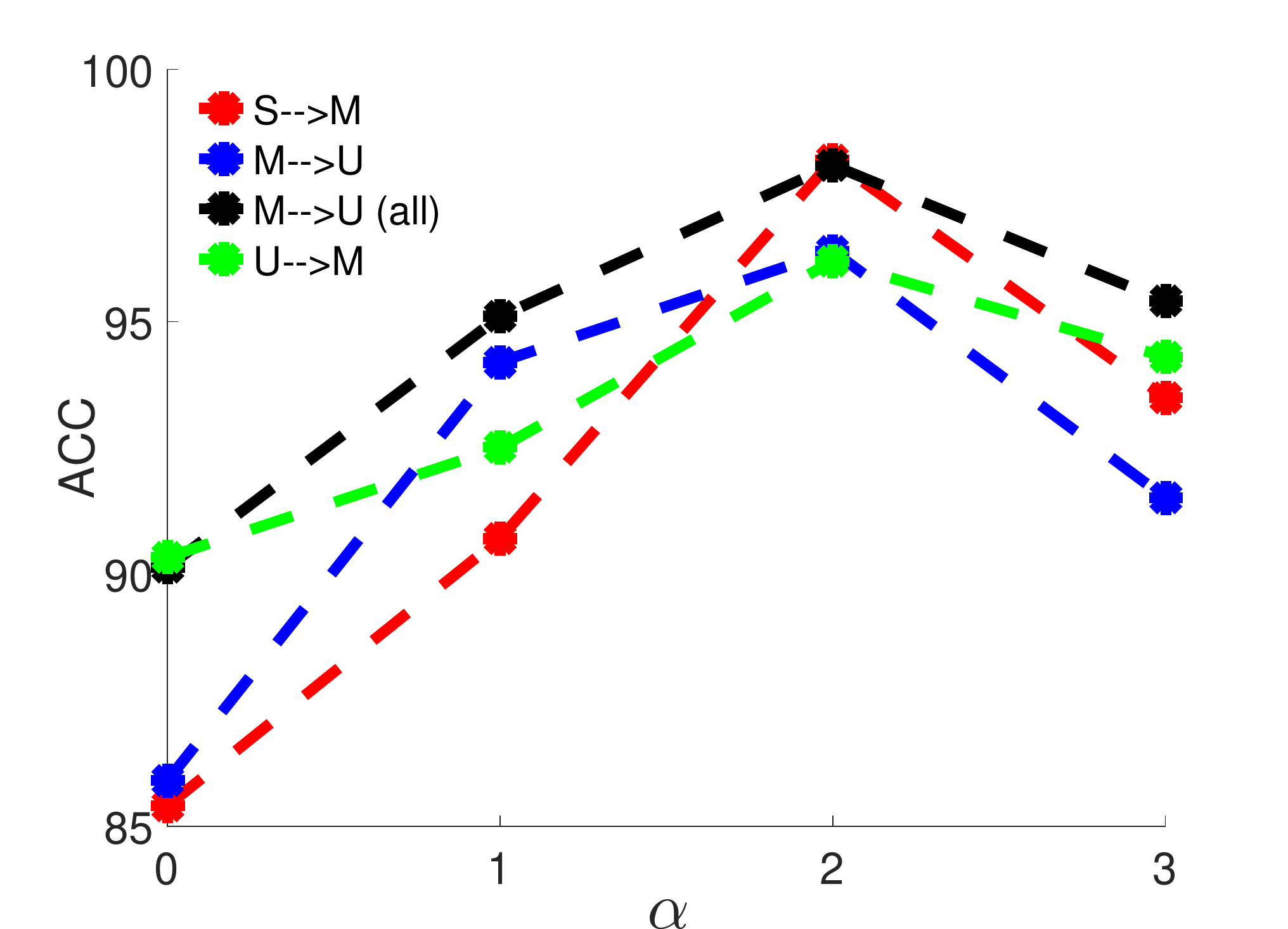}
            \caption{Sensitivity to $\alpha$\label{fig:abl_alpha}}
        \end{subfigure}
        \begin{subfigure}[t]{.45\linewidth}
         \includegraphics[trim={-2.7cm 0.45cm 0cm 0.5cm},clip, scale=0.31,valign=t 
          ]{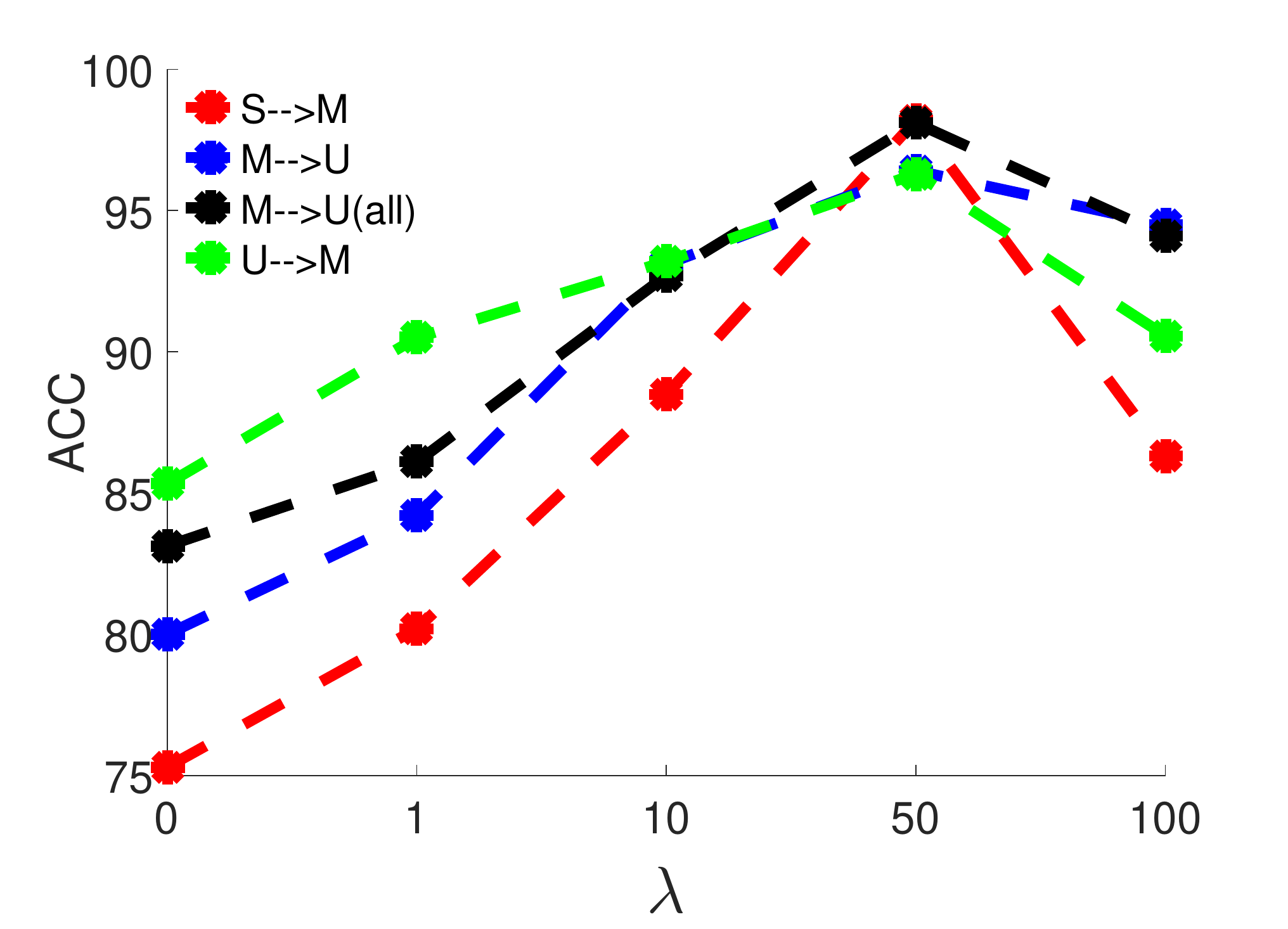}
        \caption{Sensitivity to $\lambda$\label{fig:abl_lambda}}
        \end{subfigure}
	\vspace{-0.7em}
	\caption{Sensitivity analysis of our \textbf{GPDA} on the Digit datasets. $S\to M$ denotes adaptation from \textbf{SVHN} to \textbf{MNIST} (similarly for others), and $M\to U\ (all)$ indicates using all training samples.}\label{fig:abl}
\vspace{-0.5em}
\end{figure}

Two complementary studies are conducted to investigate the impact of two hyper-parameters $\alpha$ and $\lambda$, controlling the trade off of the variance of the classifier's posterior distribution and the \textbf{MCD} loss term, respectively.
To this end, we conducted additional experiments for the digit datasets to analyze the parameter sensitivity of \textbf{GPDA} w.r.t. $\alpha$ and $\lambda$, with results depicted in \autoref{fig:abl_alpha} and~\ref{fig:abl_lambda}, respectively. 
Sensitivity analysis is performed by varying one parameter at the time over a given range, while for the other parameters we set them to their final values $(\alpha = 2, \lambda = 50)$. From \autoref{fig:abl_lambda}, we see that when $\lambda = 0$ (no \textbf{MCD} regularization term), the performance drops considerably. As $\lambda$ increases from $0$ to $50$, the performance also increases demonstrating the benefit of hypothesis consistency (\textbf{MS} term) over the target samples. Indeed, using the proposed learning scheme, we find a representation space in which we embed the knowledge from the target domain into the learned classifier.

Similarly, from \autoref{fig:abl_alpha}, we see that when $\alpha = 0$ (no prediction uncertainty) the classification accuracy is lower than the case where we utilize the prediction uncertainty, $\alpha >0$.
The key observation is that it is more beneficial to make use of the information from the full posterior distribution of the classifier during the learning process in contrast to when the classifier is considered as a deterministic function.

\subsection{Prediction Uncertainty vs.~Prediction Quality}

Another advantage of our \textbf{GPDA} model, inherited from Bayesian modeling, is that it provides a quantified measure of prediction uncertainty. In the multi-class setup considered here, this uncertainty amounts to the degree of overlap between two largest mean posteriors,  $p(f_{j^*}({\bf z})|\mathcal{D}_S)$ and $p(f_{j^\dagger}({\bf z})|\mathcal{D}_S)$, where $j^*$ and $j^\dagger$ are the indices of the largest and the second largest among the posterior means $\{\mu_j({\bf z})\}_{j=1}^K$, respectively (c.f., (\ref{eq:gp_post_mean})). Intuitively, if the two are overlapped significantly, our model's decision is less certain, meaning that we anticipate the class prediction may not be trustworthy. On the other hand, if the two are well separated, we expect high prediction quality. 

\begin{figure}[t!]
\begin{center}
\includegraphics[trim = 5mm 0mm 5mm 0mm, clip, scale=0.905]{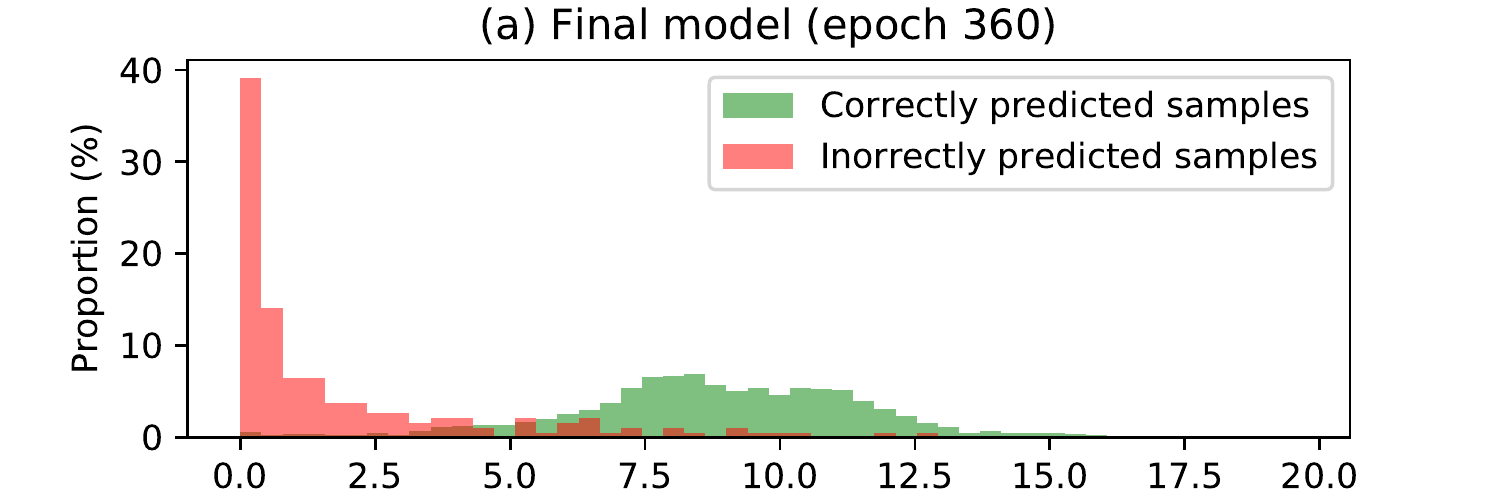} \\
\vspace{+0.2em}
\includegraphics[trim = 5mm 0mm 5mm 0mm, clip, scale=0.905]{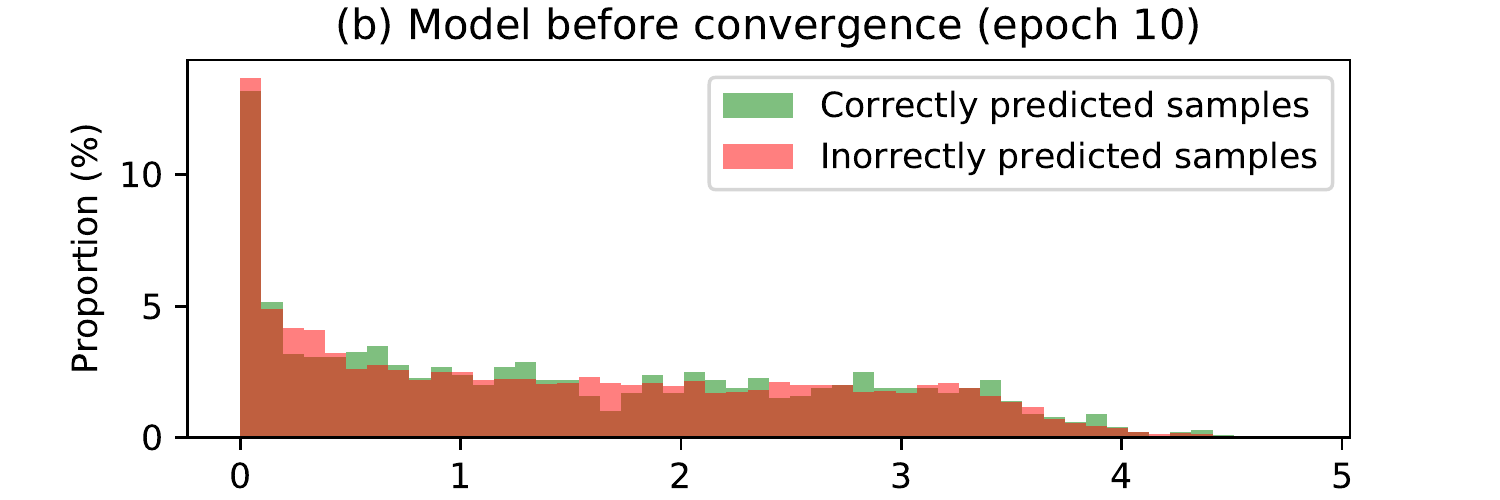}
\end{center}
\vspace{-1.7em}
\caption{Histograms of prediction (un)certainty for our two models: (a) after convergence, (b) at an early stage of training. The X-axis is the Bhattacharyya distance b/w two largest mean posteriors, an indication of {\em prediction certainty}; the higher the distance, the more certain the prediction is. For each model, we compute histograms of correctly and incorrectly predicted samples separately (by color). In our final model (a), there is a strong correlation between prediction (un)certainty (horizontal axis) and prediction correctness (color).}
\label{fig:uncertainty}
\end{figure}

To verify this hypothesis more rigorously, we evaluate the distances between two posteriors (i.e., measure of certainty in prediction) for two different cohorts: correctly classified test target samples by our model and incorrectly predicted ones. More specifically, for the \textbf{SVHN} to \textbf{MNIST} adaptation task, we evaluate the Bhattacharyya distances~\cite{derpanis2008bhattacharyya} for the samples in the two cohorts. In our variational Gaussian approximation (\ref{eq:approx_post_mean_stdev}), the Bhattacharyya distance can be computed in a closed form (See Appendix in supplementary for details). 

The histograms of the distances are depicted in \autoref{fig:uncertainty} where we contrast the two models, one at an early stage of training and the other after convergence.  Our final model in \autoref{fig:uncertainty}(a) exhibits large distances for most of the samples in the correctly predicted cohort (green), implying well separated posteriors or high certainty. For the incorrectly predicted samples (red), the distances are small implying significant overlap between the two posteriors, i.e., high uncertainty. 
On the other hand, for the model prior to convergence, \autoref{fig:uncertainty}(b), the two posteriors overlap strongly (small distances along horizontal axis) for most samples regardless of the correctness of prediction. This confirms that our algorithm enforces posterior separation by large margin during the training process. 
\begin{figure}[t!]
\vspace{-1.0em}
\begin{center}
\includegraphics[trim = 3mm 5mm 2mm 0mm, clip, scale=0.645]{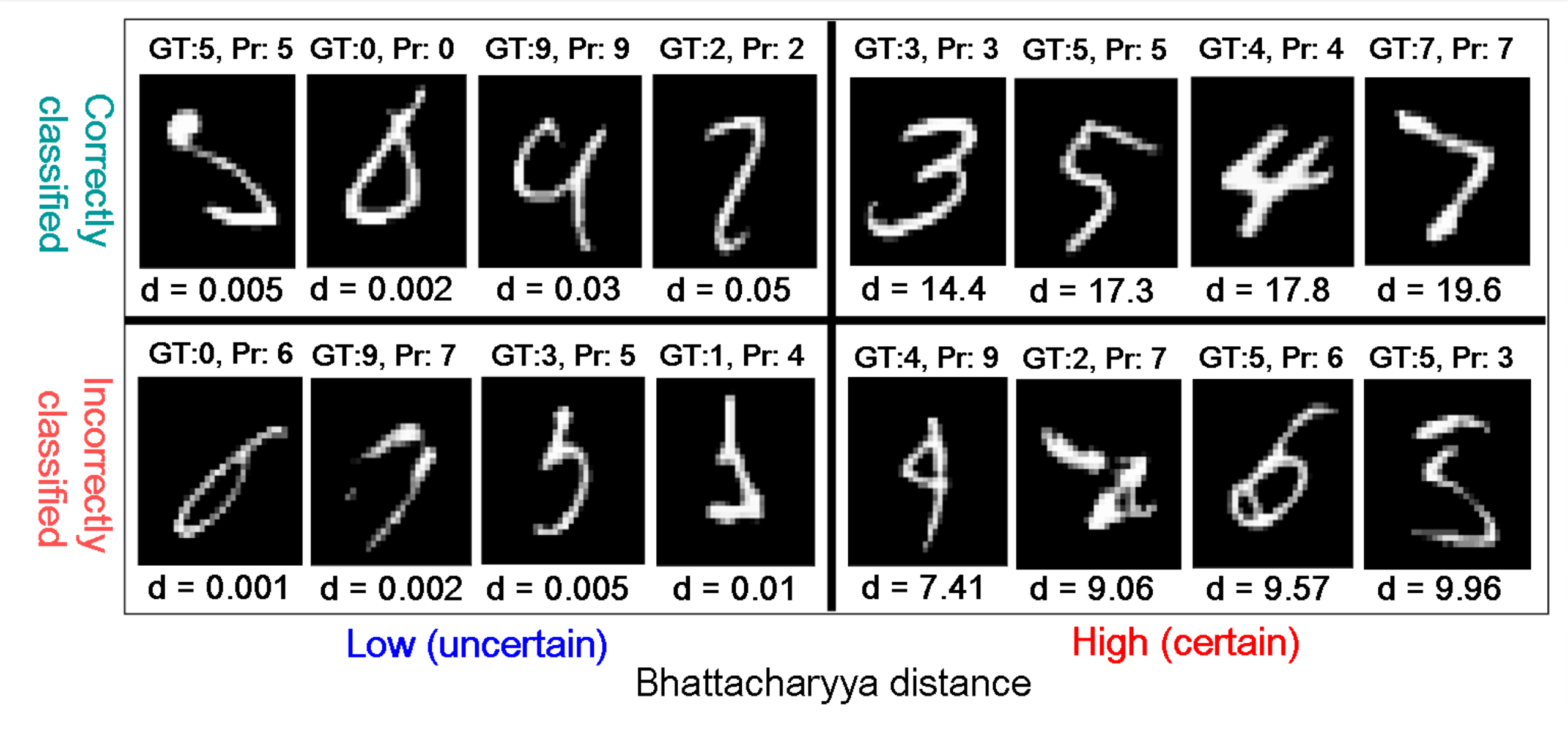}
\end{center}
\vspace{-1.5em}
\caption{Selected test (\textbf{MNIST}) images according to the Bhattacharyya distances. Right: samples with low distances (uncertain prediction). Left: high distances (certain prediction). Top: correctly classified by our model. Bottom: incorrectly classified. For each image, GT, Pr, and $d$ means ground-truth label, predicted label, and the distance, respectively.}
\label{fig:uncertainty_samples}
\end{figure}

This analysis also suggests that the measure of prediction uncertainty provided by our \textbf{GPDA} model, can be used as an {\em indicator of prediction quality}, namely whether the prediction made by our model is trustworthy or not. 
To verify this, we depict some sample test images in \autoref{fig:uncertainty_samples}. We differentiate samples according to their Bhattacharyya distances. When the prediction is uncertain (left panel), we see that the images are indeed difficult examples even for human. An interesting case is when the prediction certainty is high but incorrectly classified (lower right panel), where the images look peculiar in the sense that humans are also prone to misclassify those with considerably high certainty.

\begin{figure}[h]
\vspace{-1.3em}
\centering\setlength\tabcolsep{1.5pt}
     \begin{tabular}{cc} 
\vspace{-0.5em}
    \begin{subfigure}[b]{0.45\linewidth}
    \includegraphics[trim={0.2cm 2.0cm 0cm 0.7cm},clip,width = \linewidth]{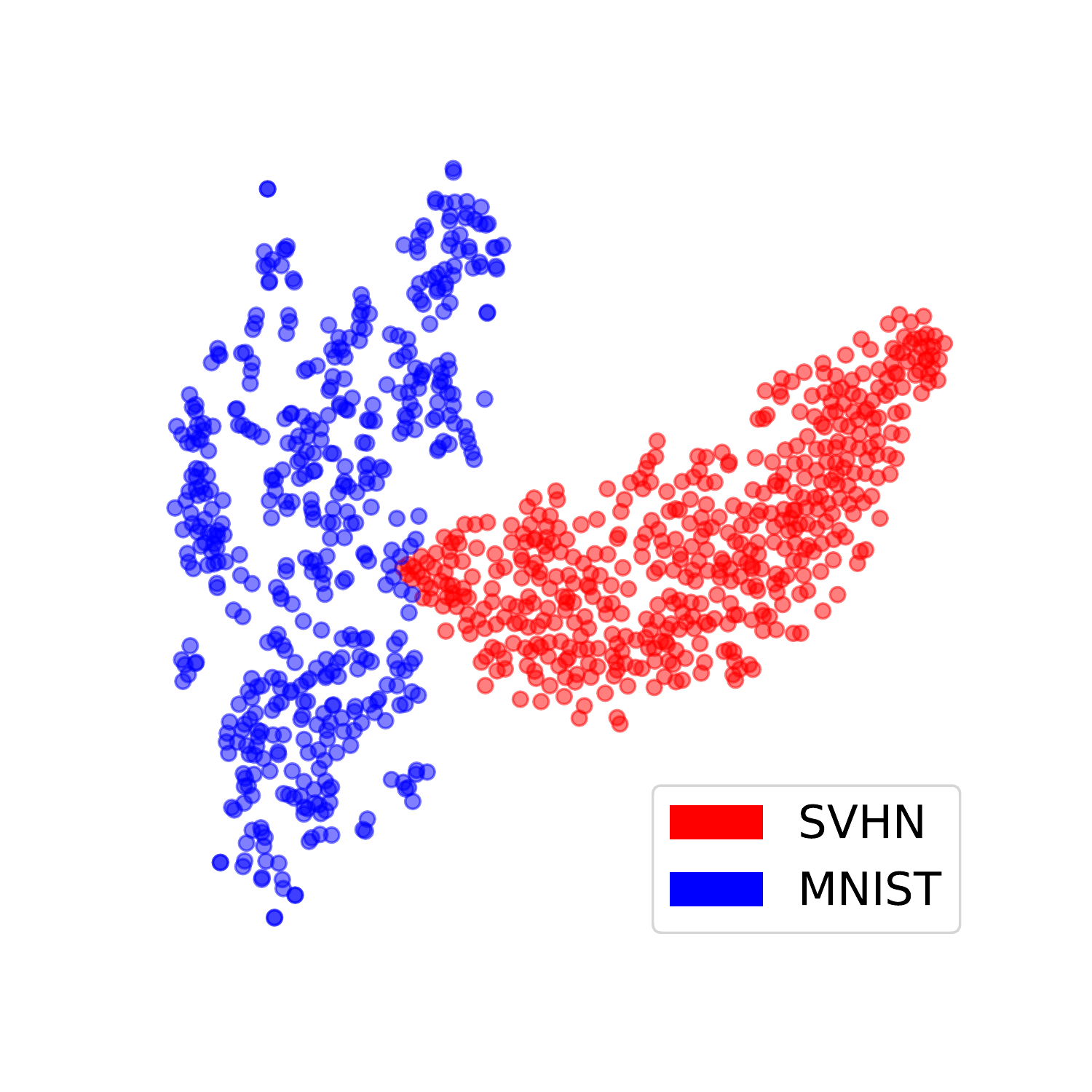}
    \caption{Original (by domain)\label{fig:embed_d_orig}}
    \end{subfigure}
    &
    \begin{subfigure}[b]{0.45\linewidth}
	 \includegraphics[trim={0.2cm 2.0cm 0cm 0.7cm},clip,width = \linewidth]{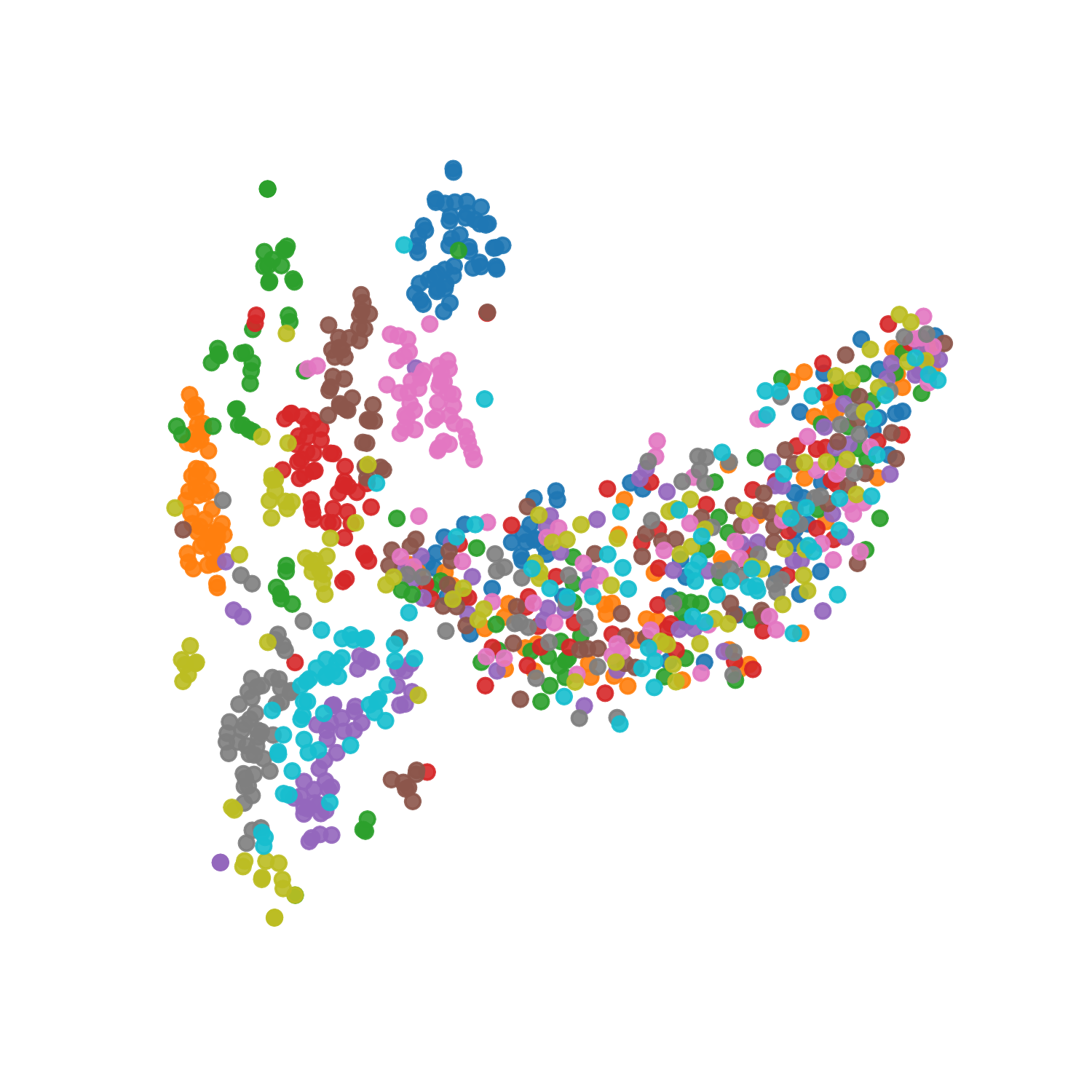}
	 \caption{Original (by classes)
	 \label{fig:embed_c_orig}}
	 \end{subfigure}
	\\
	 \vspace{-0.8em}
	 \begin{subfigure}[b]{0.45\linewidth}
       \includegraphics[trim={0.2cm 2.0cm 0cm 0.7cm},clip,width = \linewidth]{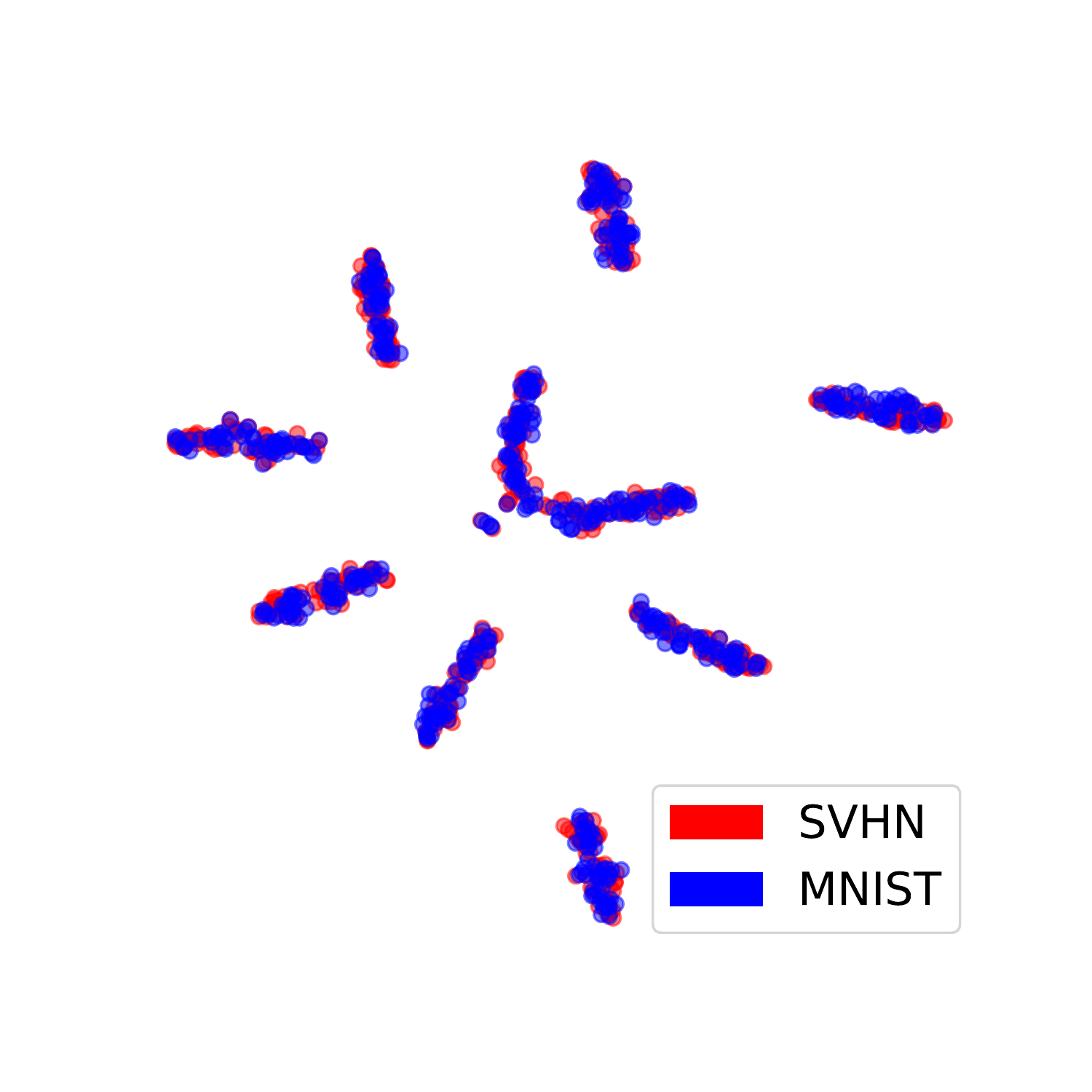}
       \caption{GPDA (by domain)\label{fig:embed_d_all}}
     \end{subfigure}
     & 
	\begin{subfigure}[b]{0.45\linewidth}
	\includegraphics[trim={0.2cm 2.0cm -0cm 0.7cm},clip,width = \linewidth
	]{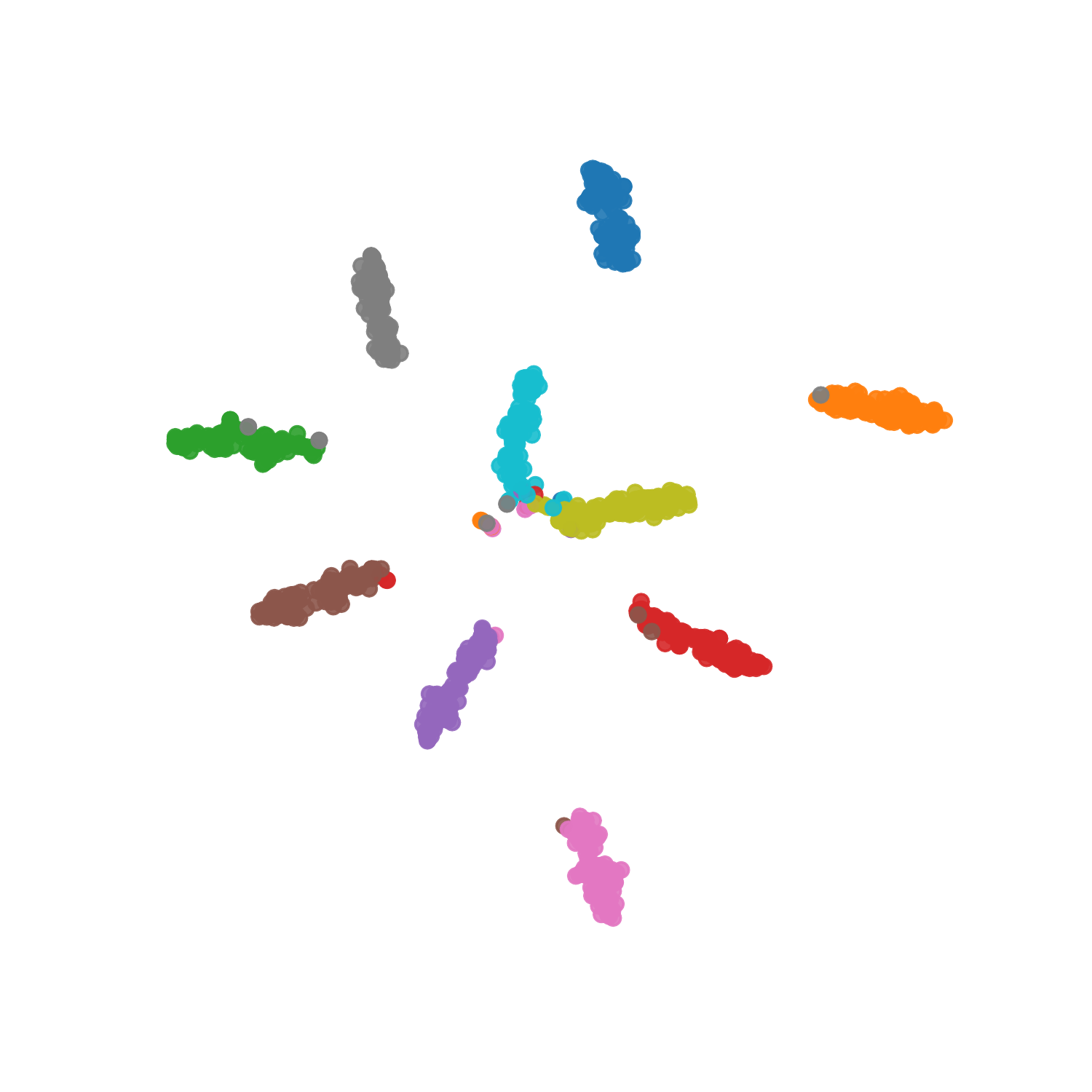}
	\caption{GPDA (by classes)\label{fig:embed_c_all}}
	\end{subfigure}
    \end{tabular}
    \caption{Feature visualization for embedding of digit datasets for adapting \textbf{SVHN} to \textbf{MNIST} using t-SNE algorithm. The first and the second columns show the domains and classes, respectively, with color indicating domain and class membership. \protect\subref{fig:embed_d_orig},\protect\subref{fig:embed_c_orig} Original features.
     \protect\subref{fig:embed_d_all},\protect\subref{fig:embed_c_all} learned features for \textbf{GPDA}.
     }
     \label{vis}
\end{figure}




\subsection{Analysis of Shared Space Embedding}
We use t-SNE~\cite{maaten2008visualizing} on \textbf{VisDA} dataset to visualize the feature representations from different classes. \autoref{vis} depicts the embedding of the learned features $G(\bf x)$, and the original features $\bf x$. Colors indicate source (red) and target (blue) domains. Notice that \textbf{GPDA} significantly reduces the domain mismatch, resulting in the expected tight clustering. This is partially due to the use of the proposed probabilistic \textbf{MCD} approach, which shrinks the classifier hypothesis class to contain only consistent classifiers on target samples while exploiting the uncertainty in the prediction.

\section{Conclusion}\label{sec:conclusion}
We proposed a novel {\it probabilistic} approach for UDA that learns an efficient domain-adaptive classifier with strong generalization to target domains. The key to the proposed approach is to model the classifier's hypothesis space in Bayesian fashion and impose consistency over the target samples in their space by constraining the classifier's posterior distribution. To tackle the intractability of computing the exact posteriors, we combined the variational Bayesian method with a deep kernel technique to efficiently approximate the classifier's posterior distribution. We showed on three challenging benchmark datasets for image classification that the proposed method outperforms current state-of-the-art in unsupervised domain adaptation of visual categories.

\appendix

\newpage

\begin{center}
\LARGE Supplementary Material
\end{center}

\section{Overview}
In this Supplement, we present additional analyses highlighting the ability of our model, GPDA, to leverage its inherent measure of uncertainty to both produce increasingly accurate predictions as well as provide a measure of its own trustworthiness. These new results are summarized in \autoref{sec:uncertainty}.  \autoref{sec:remark_max_margin} provides further analysis showing the key connection between GPDA and the max-margin Gaussian Process classification in the original space $\mathcal{X}$, surpassing the explicit need for a shared space $\mathcal{Z}$ of traditional domain adaptation approaches.  We then present specific details of all datasets used in our experiments as well as the particulars of relevant experimental setups in \autoref{sec:datasets}.  Finally, we provide a brief overview of Gaussian Process models in \autoref{sec:gpmodel} and another related state-of-the-art domain adaptation approach, the MCDA, in \autoref{sec:mcd}.

\section{Additional Analyses: Prediction Uncertainty vs.~Prediction Quality}\label{sec:uncertainty}

A key benefit of our GPDA algorithm, inherited from Bayesian modeling, is that it provides a quantified measure of prediction uncertainty. In the multi-class setup, for an input ${\bf x}$ we  measure the uncertainty as the degree of overlap between the two largest mean posteriors,  $p(f_{j^*}({\bf z})|\mathcal{D}_S)$ and $p(f_{j^\dagger}({\bf z})|\mathcal{D}_S)$, where ${\bf z} = {\bf G}({\bf x})$, $j^*$ and $j^\dagger$ are the indices of the largest and the second largest among the posterior means $\{\mu_j({\bf z})\}_{j=1}^K$, respectively, 
If the two overlap significantly, our model's decision is less certain, signifying that we anticipate the class prediction not to be trustworthy. On the other hand, if the two are well separated, we expect high prediction quality.\\

\noindent\textbf{Bhattacharyya distance.} In the main paper (Sec.~5.4 and Fig.~4), we have verified this hypothesis by evaluating the Bhattacharyya distances (BD) between two posteriors (i.e., a measure of {\em certainty} in prediction) for two different cohorts: correctly classified test target samples by our model and incorrectly predicted ones, for the \textbf{SVHN} to \textbf{MNIST} adaptation task.  
Since we use variational Gaussian approximation of the posteriors $p(f_j({\bf z})|\mathcal{D}_S) \approx \mathcal{N}(\mu_j({\bf z}),\sigma_j({\bf z})^2)$, 
where $\mu_j({\bf z})$ and $\sigma_j({\bf z})$ are determined by Eq.~(22) in the main paper, the Bhattacharyya distance can be computed in closed form:
\begin{equation}
\textrm{BD} = \frac{1}{4} \log \left ( \frac 1 4 \left( \frac{\sigma_{j^*}^2}{\sigma_{j^\dagger}^2}+\frac{\sigma_{j^\dagger}^2}{\sigma_{j^*}^2}+2\right ) \right ) +\frac{1}{4} \left ( \frac{(\mu_{j^*}-\mu_{j^\dagger})^{2}}{\sigma_{j^*}^2+\sigma_{j^\dagger}^2}\right ).
\end{equation}

\begin{figure}
\begin{center}
\includegraphics[trim = 5mm 0mm 5mm 0mm, clip, scale=0.605]{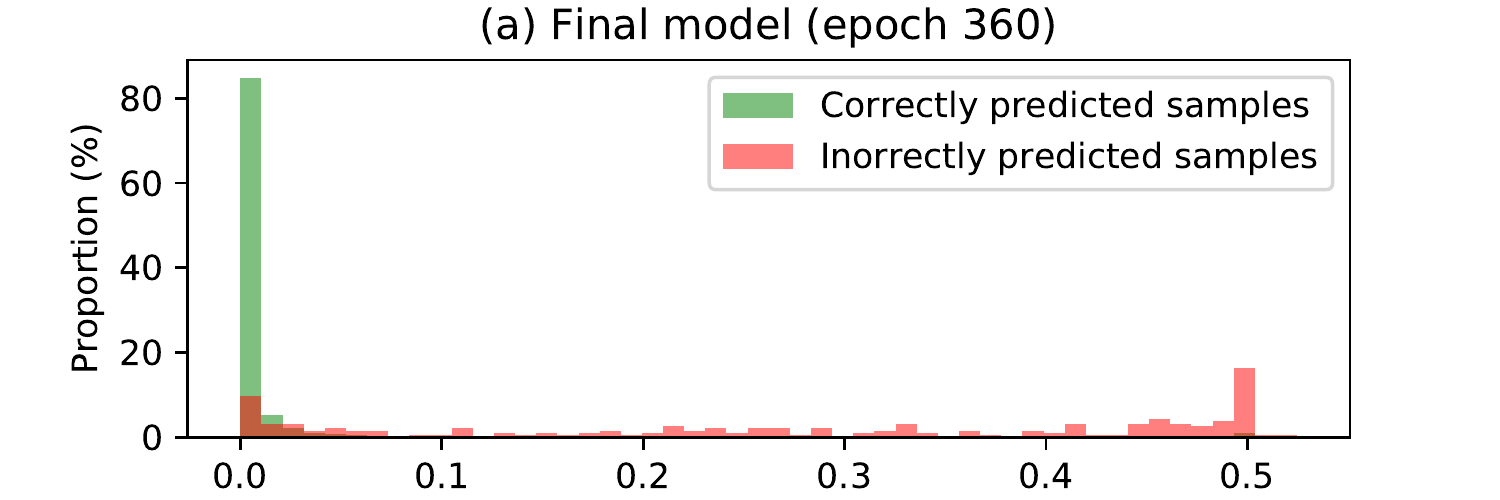} 
\includegraphics[trim = 5mm 0mm 5mm 0mm, clip, scale=0.605]{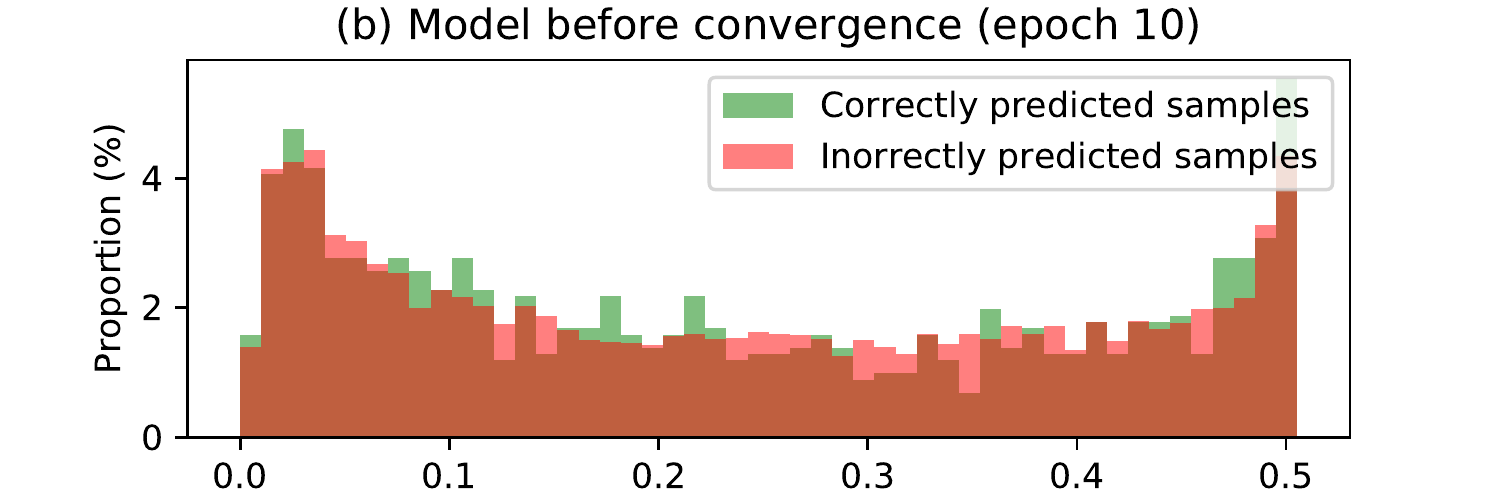}
\end{center}
\vspace{-1.7em}
\caption{(For our GPDA) Histograms of Bayes error rates (prediction uncertainty) for our two models: (a) after convergence, (b) at an early stage of training. The X-axis is the Bayes optimal error rate~(\ref{eq:bayes_error}) b/w two largest mean posteriors, an indication of {\em prediction uncertainty}; the higher the error rate, the more uncertain the prediction is. For each model, we compute histograms of correctly and incorrectly predicted samples separately (by color). In our final model (a), there is a strong correlation between prediction uncertainty (horizontal axis) and prediction correctness (color).}
\label{fig:uncertainty_bayes_error}
\end{figure}

\noindent\textbf{Bayes Optimal Error Rate.} An alternative metric to measure the prediction uncertainty, perhaps more principled in the Bayesian sense, is the Bayes optimal error rate between the two largest mean posteriors, which can be computed as
\begin{equation}
\textrm{Bayes error} = \frac{1}{2} \int_{D}^{\infty} \mathcal{N}(x; \mu_{j^\dagger},\sigma_{j^\dagger}^2) \ dx  + \frac{1}{2} \int_{-\infty}^{D} \mathcal{N}(x; \mu_{j^*},\sigma_{j^*}^2) \ dx  = \frac{1}{2} \Bigg( \Phi\bigg( \frac{\mu_{j^\dagger}-D}{\sigma_{j^\dagger}}\bigg) + \Phi\bigg( \frac{D-\mu_{j^*}}{\sigma_{j^*}}\bigg) \Bigg),
\label{eq:bayes_error}
\end{equation}
where $\Phi$ is the CDF of $\mathcal{N}(0,1)$ and $D$ is the Bayes optimal decision threshold, $D = (\mu_{j^*} - \mu_{j^\dagger}) /  \sqrt{ (\sigma_{j^*}^2 + \sigma_{j^\dagger}^2)/2 }$. 
The interpretation is: the smaller the Bayes error rate, the more certain our prediction is, and vice versa. We depict the histograms of the Bayes error rates for two cohorts in \autoref{fig:uncertainty_bayes_error}. 
As shown, the conclusion is very similar to our earlier analysis based on Bhattacharyya distances: Our final model in \autoref{fig:uncertainty_bayes_error}(a) exhibits low error rates for most of the samples in the correctly predicted cohort (green), implying well separated posteriors or high certainty of prediction. For the incorrectly predicted samples (red), the error rates are mostly high implying significant overlap between the two posteriors, i.e., high uncertainty of prediction. \\

\noindent\textbf{Uncertainty in GPDA vs MCDA.}  Lastly, to demonstrate that it is the unique property of our GPDA model that the uncertainty measure can be used to credibly gauge the quality of prediction at test time, we contrast our model with other non-Bayesian approaches. Specifically, we consider MCDA, as the second-best competing method. The MCDA is a non-Bayesian method that yields {\em point estimate} class prediction, namely $p(y|{\bf x})$. 
By point estimate, we mean that the MCDA prediction places all its probability mass on a single (softmax) probability (score) value $p(y=j|{\bf x})$ for each class $j$, unable to provide a degree of uncertainty in its prediction (e.g., $\sigma_j$ in our GPDA model). 

However, one can define a {\em heuristic} notion of uncertainty for the MCDA by measuring how distant the two largest score predictions are from each other. More specifically, we compute the following quantity, dubbed Bhattacharyya {\em  pseudo} distance (BPD), as a measure of uncertainty in the MCDA:
\begin{equation}
\textrm{BPD} := \log  p(y=j^* | {\bf x}) - \log p(y=j^\dagger | {\bf x})
\label{eq:pseudo_bhatt}
\end{equation}
where $j^*$ and $j^\dagger$ are the indices of the largest and the second largest among the scores $\log p(y=j|{\bf x})$, respectively. 
Note that (\ref{eq:pseudo_bhatt}) is the log-ratio between the largest two class prediction scores. We name it the {\em  pseudo} distance as it reduces to the Bhattacharyya distance if we form Gaussians with the mean equal to $\log p(y=j|{\bf x})$ and the same variances for both $j^*$ and $j^\dagger$.

\begin{figure}
\begin{center}
\includegraphics[trim = 5mm 0mm 5mm 0mm, clip, scale=0.605]{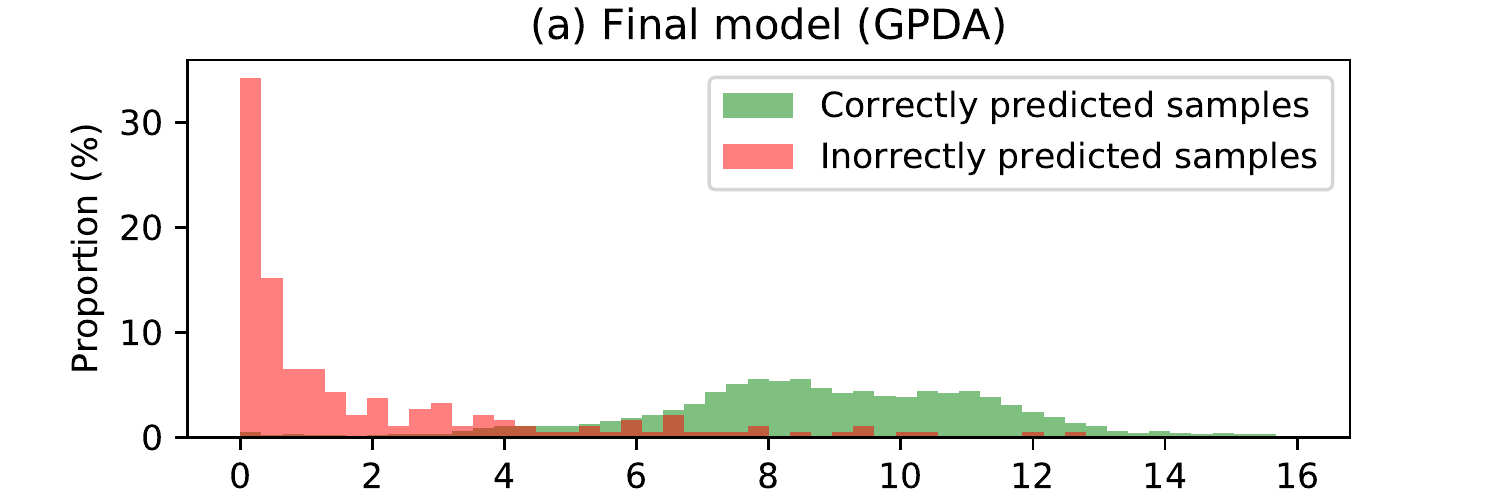} 
\includegraphics[trim = 5mm 0mm 5mm 0mm, clip, scale=0.605]{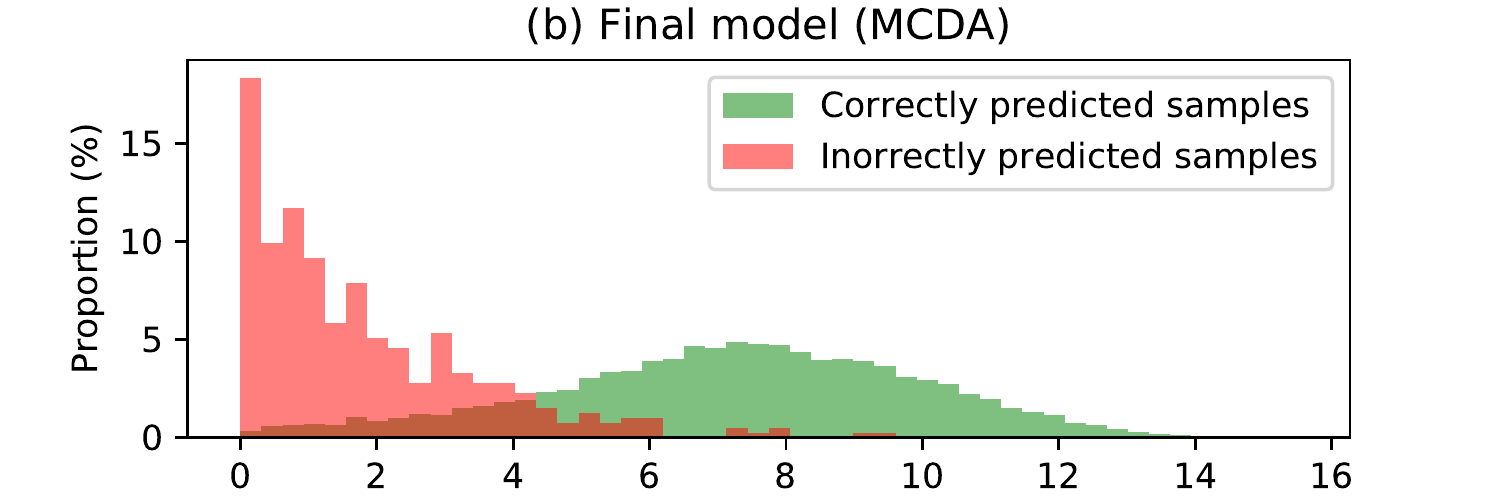}
\end{center}
\vspace{-1.7em}
\caption{(GPDA vs.~MCDA) Histograms of Bhattacharyya distances between two largest mean posteriors (prediction certainty) for (a) GPDA and (b) MCDA. The X-axis is the Bhattacharyya distance, an indication of {\em prediction certainty}; the higher the distance, the more certain the prediction is. For the non-Bayesian point-estimate-based MCDA, we compute the  Bhattacharyya pseudo distance instead, as described in the text. Qualitatively, our GPDA model exhibits stronger correlation (histograms less overlapped) between prediction uncertainty (horizontal axis) and prediction correctness (color).}
\label{fig:uncertainty_bhatt_both}
\end{figure}

We depict the histograms of the pseudo distances for MCDA's two cohorts in \autoref{fig:uncertainty_bhatt_both}(b), where the Bhattacharyya histograms for our GPDA are also shown in \autoref{fig:uncertainty_bhatt_both}(a) for comparison. Unlike the more clear separation attained in our GPDA model, the MCDA exhibits two issues: i) For the correctly predicted samples (green), a considerable number of points have large overlap between $j^*$ and $j^\dagger$ (i.e., low BPDs).
ii) For the incorrectly predicted samples (red), the number of cases where the two largest scores are relatively well separated\footnote{E.g., those with $\textrm{BPD} > 1.0$, namely, certain predictions.} exceeds that of our GPDA model, suggesting higher prediction uncertainty.
This signifies the unique benefit of our Bayesian domain adaptation approach, that is, the \textit{capability to utilize the prediction uncertainty as a gauge of prediction quality}.\\

\begin{figure}
\vspace{-1.0em}
\begin{center}
\includegraphics[trim = 3mm 5mm 2mm 2mm, clip, scale=0.338]{uncertainty_quality_samples_new}
\includegraphics[trim = 3mm 5mm 2mm 2mm, clip, scale=0.338]{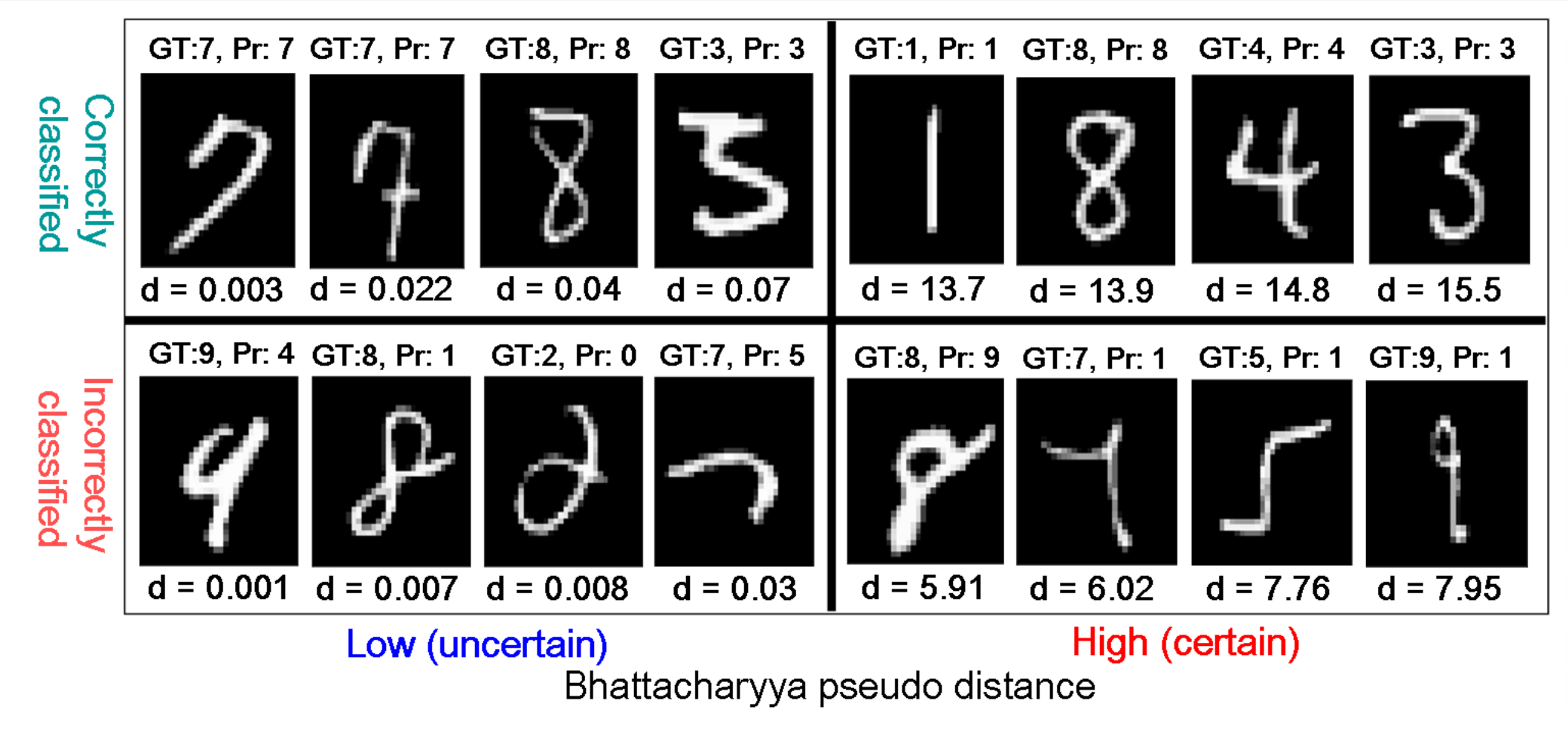}\\
\ \ \ \ (a) GPDA \ \ \ \ \ \ \ \ \ \ \ \ \ \ \ \ \ \ \ \ \ \ \ \ \ \ \ \ \ \ \ \ \ \ \ \ \ \ \ \ \ \ \ \ \ \ \ \ \ \ \ \ \ \ \ \ \ \ \ \ \ \ \ \ \ \ \ \ \ \ \ \ \ \ \ \ \ \ \ \ \ (b) MCDA
\end{center}
\vspace{-1.5em}
\caption{Selected test (\textbf{MNIST}) images according to the Bhattacharyya (pseudo) distances estimated by (a) GPDA and (b) MCDA. For each figure, Left: samples with low distances (uncertain prediction). Right: high distances (certain prediction). Top: correctly classified by the model. Bottom: incorrectly classified. For each image, GT, Pr, and $d$ stand for ground-truth label, predicted label, and the (pseudo) distance, respectively.}
\label{fig:uncertainty_samples_both}
\end{figure}

\noindent\textbf{GPDA vs. MCDA -- Hard vs. Easy Instances.} As a counterpart to Fig.~5 in the main paper, we also depict in \autoref{fig:uncertainty_samples_both}(b) some sample target test images that are correctly/incorrectly predicted by the MCDA with low/high certainty according to the BPD. For ease of comparison, we also show the samples for our GPDA from the main paper, Fig. 5, in \autoref{fig:uncertainty_samples_both}(a). 
Unlike the GPDA, the uncertainty prediction made by the MCDA shows less agreement with the human assessment. Images whose BPDs are low (i.e., uncertain prediction judged by the MCDA shown in the left panel of \autoref{fig:uncertainty_samples_both}(b)), appear to be visually easy to classify by a human, with no ambiguity, with a possible exception in few cases: e.g., the last example in the correct/low quadrant that may look like "five", while the first example in the incorrect/low quadrant may be interpreted as "four".  Furthermore, the sample images in the incorrect/high quadrant of \autoref{fig:uncertainty_samples_both}(b), i.e., those predicted by the MCDA with high certainty but misclassified, are relatively easy-to-classify examples for a human, other than the second example that may be confused as "one". 

This empirical analysis verifies that the measure of prediction uncertainty provided by our GPDA model can be used as a more accurate indicator of prediction quality than that implied by the MCDA, our top competitor. That is, our model's quantitative uncertainty measure can determine, with high precision, whether the prediction made by the model is trustworthy or not.

\section{A Remark on Proposed GPDA Algorithm}\label{sec:remark_max_margin}

In this section we discuss the strong connection between the GPDA algorithm 
and the max-margin confident prediction (or the entropy minimization) framework in classical semi-supervised learning~\cite{ssem04,semisup_book}. More specifically, we show that our GPDA algorithm, in the algorithmic point of view, can be viewed as a {\em max-margin Gaussian process classifier} on the original input space $\mathcal{X}$ without explicitly considering a shared space $\mathcal{Z}$. 

Recall that the GPDA algorithm can be summarized as the following two alternating optimizations: 
\begin{mdframed}
\begin{itemize}
\item $\min_{\{{\bf m}_j,{\bf S}_j\}} \  -\textrm{LL} + \textrm{KL}$ \ \ \ \ \ 
(variational inference)
\item $\min_{{\bf G},k} 
\  -\textrm{LL} + \textrm{KL} + \lambda \cdot \textrm{MS}$ \ \ \ \ (model selection)
\end{itemize}
\end{mdframed}
where the key terms in these objectives are defined as follows:
\begin{equation}
\textrm{KL} = \frac{1}{2} \sum_{j=1}^K \big( \ 
  \textrm{Tr}({\bf S}_j) + ||{\bf m}_j||_2^2 - \log\det({\bf S}_j) - d \ \big),
\label{eq:kl} 
\end{equation}
\begin{equation}
\textrm{LL} = \frac{1}{M} \sum_{m=1}^M \frac{N_S}{|B_S|} \sum_{i \in B_S}
    \log P(y^S_i | {\bf W}^{(m)} {\boldsymbol \phi}({\bf z}^S_i)),
\label{eq:ll} 
\end{equation}
and
\begin{equation}
\textrm{MS} := \frac{1}{|B_T|} \sum_{i\in B_T} \bigg( 
  \max_{j \neq j^*} {\bf m}_j^\top {\boldsymbol \phi}({\bf z}^T_i) - 
  \max_{1\leq j \leq K} {\bf m}_j^\top {\boldsymbol \phi}({\bf z}^T_i) \ + \ 1 \ + \ \alpha \max_{1\leq j \leq K} 
    \big( {\boldsymbol \phi}({\bf z}^T_i)^\top {\bf S}_j {\boldsymbol \phi}({\bf z}^T_i) \big)^{1/2} 
\bigg)_+.
\label{eq:ms}
\end{equation}
Note that ${\bf z} = {\bf G}({\bf x})$. 
Although we have built a \textbf {GP} classification model on top of the shared space $\mathcal{Z}$, leading to the algorithm above,  in our learning objective terms  (\ref{eq:kl}--\ref{eq:ms}), the deep kernel feature mapping ${\boldsymbol\phi}(\cdot)$ and the embedding function ${\bf G}(\cdot)$ always appear together in the composite form ${\boldsymbol\phi}({\bf G}(\cdot))$. 

Thus, our approach is functionally equivalent to building a \textbf {GP} classification model on top of the original $\mathcal{X}$ space, where the explicit feature mapping is ${\bf x} \to ({\boldsymbol \phi} \circ {\bf G})({\bf x})$. More formally, our classifier can be written as ${\bf f}({\bf x}) 
= {\bf W} {\boldsymbol \phi}({\bf G}({\bf x}))$, a function of ${\bf x}$.
Consequently, our approach can be viewed as a {\em max-margin Gaussian process classifier}, without explicitly considering the shared space, where we push the posterior inferred from the source domain data to meet the large margin criterion on the (unlabeled) target domain data. This is clearly in line with {\em entropy minimization} or {\em max-margin confident prediction} principles in classical semi-supervised learning~\cite{ssem04,semisup_book}.

\section{Details of Datasets and Experimental Setups}\label{sec:datasets}
We now present additional details of experiments on the three datasets used in the main paper. For all experiments, we set $M=50$, the number of posterior samples from the variational density $q({\bf W})$ (Sec. 3.2 in the main paper for more details). 
\subsection{Digit and Traffic Signs Datasets}
We followed the experimental setup used in~\cite{ganin2014unsupervised} in the following three adaptation scenarios. For this experiment, we compare our GPDA model with various state-of-the-art unsupervised domain adaptation approaces, namely: \textbf{MMD}~\cite{long2015learning}, \textbf{DANN}~\cite{ganin2014unsupervised}, \textbf{DSN}~\cite{bousmalis2016domain}, \textbf{ADDA}~\cite{tzeng2017adversarial},
\textbf{CoGAN}~\cite{NIPS2016_6544},
\textbf{PixelDA}~\cite{bousmalis2017unsupervised},
\textbf{ATDA}~\cite{saito2017asymmetric},
\textbf{ASSC}~\cite{haeusser2017associative},
\textbf{DRCN}~\cite{ghifary2016deep},
and, \textbf{MCDA}~\cite{saito2018}.\\

\begin{itemize}
\item \textbf{SVHN$\rightarrow$MNIST.}
In this adaptation scenario, we used the standard training set as our training samples, and the standard testing set as our testing samples both for source and target samples.

\item \textbf{SYN SIGNS$\rightarrow$GTSRB.}
Following \textbf{MCDA}~\cite{saito2018}, we randomly selected 31367 samples for the target training set and evaluated the accuracy on the remaining samples. 

\item \textbf{MNIST$\leftrightarrow$USPS.}
For this experiment, we followed the protocols used in ADDA~\cite{tzeng2017adversarial} and PixelDA~\cite{bousmalis2017unsupervised}. ADDA provides the setting where a part of training samples are utilized during training. 2,000 training samples are picked up for MNIST and 1,800 samples are used for USPS.
PixelDA allows one to utilize all of the standard training samples during learning.
\end{itemize}

\subsection{VisDA Dataset}
We used VisDA dataset~\cite{peng2017visda}  to evaluate adaptation from synthetic to real-object images. The dataset is an instance of cross-domain object classification, with over 280K images across 12 categories in the combined training, validation, and testing domains. The source images, 152,397 synthetic images, were generated by rendering 3D models of the same object categories as in the real data from different angles and under different lighting conditions. The validation set of 55,388 images was collected from MSCOCO~\cite{lin2014microsoft}. In our experiment, we considered the images of validation splits as the target domain and trained models in the unsupervised domain adaptation settings. We evaluate the performance of ResNet101~\cite{he2016deep} model pre-trained on Imagenet~\cite{deng2009imagenet}. For this experiment, we compare our model with \textbf{MMD}~\cite{long2015learning}, \textbf{DANN}~\cite{ganin2014unsupervised}, and \textbf{MCDA}~\cite{saito2018}.

\section{Background -- Gaussian Process}\label{sec:gpmodel}

A Gaussian Process (GP) is an infinite collection of random variables $\{f(\mathbf{x})| \mathbf{x} \in X \}$, such that any finite number of samples have a joint Gaussian distribution. A GP is fully specified by the mean function $\mu(\mathbf{x})$ and the covariance function $k(\mathbf{x}, \mathbf{x'})$, typically user-defined. GPs can also be interpreted as a distribution over functions $f(\mathbf{x}) \sim \mathcal{GP}(\mu(x), k(x,x))$ such that any finite collection of function values $[f(\mathbf{x_{1}}), \ldots, f(\mathbf{x_{N}})]$ have a joint Gaussian distribution: 
\begin{equation}
[f(\mathbf{x_{1}}), \ldots, f(\mathbf{x_{N}})] \sim \mathcal{N}(\bm{\mu},K),
\end{equation}
where $\bm{\mu}$ is the $N \times 1$ vector $\mu_{i} = \mu(\mathbf{x_{i}})$ and $K$ is the $N \times N$ covariance matrix with $K_{ij} = k(\mathbf{x_{i}, \mathbf{x_{j}}})$. 

A training dataset consists of $N$ pairs of data $(\mathbf{x_{i}}, y_{i})_{i=1}^{N}$,  where $y_{i}$ are noisy observations of some latent function $f$ with Gaussian noise $y_{i} = f(\mathbf{x}_{i}) + \epsilon_{i}$, $\epsilon_{i} \in \mathcal{N}(0, \sigma^{2})$. The likelihood of the data $\mathbf{y}|\mathbf{f} \sim  \mathcal{N}(f, \sigma^{2}I)$ and the prior $\mathbf{f} \sim \mathcal{N}(0, K)$ give the joint probability model $p(\mathbf{f}, \mathbf{y}) = p(\mathbf{y}|\mathbf{f})p(\mathbf{f})$, where $\mathbf{y}$ denotes the noisy targets and $\mathbf{f}$ denotes the vector of underlying latent function values. The predictive distribution at a set of test points $X_{*}$ is given in closed form using the properties of conditional Gaussians, 
\begin{eqnarray}
\bf{f}_{*}|\mathbf{y}, X, X_{*}, \bm{\theta}, \sigma^{2} &\sim& \mathcal{N}(\overline{\bf{f}_{*}}, \textrm{Cov}(\bf{f}_{*})) \\
\overline{\bf{f}_{*}} &=& K_{*}(K + \sigma^2I)^{-1}\bf{y} \\
\textrm{Cov}(\bf{f}_{*}) &=& K_{**} -  K_{*}(K + \sigma^{2}I)^{-1}K_{*}^{T},
\label{pred}
\end{eqnarray}
where $K_{**}$ denotes the covariance matrix evaluated among the test inputs $X_{*}$ and $K_{*}$ denotes the covariance matrix evaluated between the test points $X_{*}$ and the training set $X$. If there are $N_{*}$ test points, the covariance matrix $K_{**}$ is of size $N_{*} \times N_{*}$ and $K_{*}$ is of size $N_{*} \times N$. 

\subsection{Gaussian Process Classification}
In Gaussian Process Classification (GPC), the target values are discrete class labels, hence it is not appropriate to model them via a multivariate Gaussian density. Instead, we use the Gaussian process as a latent function whose sign determines the class label for binary classification; for multi-class classification one can use multiple GPs or a multivariate GP.\\

The key difference between the GP regression and GPC is how the output data, $\mathbf{y}$, are connected to the underlying function
values, $\mathbf{f}$.  Precisely, they are no longer connected via a simple noise process as in the previous section, instead now discrete: for example, for binary classification framework, say $y=1$ for one class and $y=-1$ for the other. In this case, one could try fitting a GP that produces an output of $1$ for some values of $x$ and $-1$ for
others, simulating the discrete nature of the problem. Then, the classification of a new data point $x_*$ involves two steps:
\begin{enumerate}
\item Evaluate a `latent function' $f$ which models qualitatively how the likelihood of one class versus the other
changes over the $x$ axis.  This is the usual GP.
\item Squeeze the output of this latent function onto $[0,1]$ using logistic function, $\pi(f) = \sigma(y=1|f)$.
\end{enumerate}
Writing these two steps schematically,
\begin{center}
\framebox{data, $x_*$} $\xrightarrow{\text{GP}}$ \framebox{latent function, $f_*|x_*$}
$\xrightarrow{\text{sigmoid}}$ \framebox{class probability, $\pi(f_*)$}.
\end{center}

\section{Background -- MCDA~\cite{saito2018}}\label{sec:mcd}

For multi-class classification, the \textbf{MCDA} adopts classifier networks that output class prediction probabilities, $h({\bf z}) = [ p(y=1|{\bf z}), \dots, p(y=K|{\bf z}) ]^\top$.  The discrepancy between $h$ and $h'$ is defined as the expected normalized $L_1$ difference, that is, $\mathbb{E} || h({\bf z}) - h'({\bf z}) ||_1 / K$.
The learning algorithm is a coordinate descent optimization alternating among three steps:
\begin{enumerate}
\item $\min_{G,h,h'} L_S := \mathbb{E}_{({\bf x},y)\sim S} \big[ \ 
  CE(y; h({\bf G}({\bf x}))) \ + \ CE(y; h'({\bf G}({\bf x}))) \ \big]$
\item (Fix $G$) $\min_{h,h'} L_S - L_{adv}$, 
  where $L_{adv} := \mathbb{E}_{{\bf x}\sim T} \big[ \ 
    || h({\bf G}({\bf x})) - h'({\bf G}({\bf x})) ||_1 / K \ \big]$
\item (Fix $h,h'$) $\min_{G} L_{adv}$
\end{enumerate}
Here, $CE(y; p)$ stands for the cross entropy (or log) loss, i.e., $CE(y; p) = -\log p(y)$. All the expectations are approximately estimated on a mini-batch. Optionally, Step-3 can be repeated $2\sim 4$ times (on the same mini-batch) to boost the convergence of the embedding network ${\bf G}(\cdot)$.


\newpage


\begin{thebibliography}{10}\itemsep=-1pt

\bibitem{baktashmotlagh2013unsupervised}
M.~Baktashmotlagh, M.~T. Harandi, B.~C. Lovell, and M.~Salzmann.
\newblock Unsupervised domain adaptation by domain invariant projection.
\newblock In {\em IEEE International Conference on Computer Vision (ICCV)},
  pages 769--776. IEEE, 2013.

\bibitem{ben-david-2010}
S.~Ben-David, J.~Blitzer, K.~Crammer, A.~Kulesza, F.~Pereira, and J.~W.
  Vaughan.
\newblock A theory of learning from different domains.
\newblock {\em Machine Learning}, 79(1--2):151--175, 2010.

\bibitem{ben-david-2007}
S.~Ben-David, J.~Blitzer, K.~Crammer, and F.~Pereira.
\newblock Analysis of representations for domain adaptation, 2007.
\newblock In Advances in Neural Information Processing Systems.

\bibitem{benaim2017one}
S.~Benaim and L.~Wolf.
\newblock One-sided unsupervised domain mapping.
\newblock In {\em Advances in Neural Information Processing Systems (NIPS)},
  pages 752--762, 2017.

\bibitem{bengio2013representation}
Y.~Bengio, A.~Courville, and P.~Vincent.
\newblock Representation learning: A review and new perspectives.
\newblock {\em IEEE transactions on pattern analysis and machine intelligence},
  35(8):1798--1828, 2013.

\bibitem{bousmalis2017unsupervised}
K.~Bousmalis, N.~Silberman, D.~Dohan, D.~Erhan, and D.~Krishnan.
\newblock Unsupervised pixel-level domain adaptation with generative
  adversarial networks.
\newblock In {\em IEEE Conference on Computer Vision and Pattern Recognition
  (CVPR)}, volume~1, page~7, 2017.

\bibitem{bousmalis2016domain}
K.~Bousmalis, G.~Trigeorgis, N.~Silberman, D.~Krishnan, and D.~Erhan.
\newblock Domain separation networks.
\newblock In {\em Advances in Neural Information Processing Systems (NIPS)},
  pages 343--351, 2016.

\bibitem{courty2017joint}
N.~Courty, R.~Flamary, A.~Habrard, and A.~Rakotomamonjy.
\newblock Joint distribution optimal transportation for domain adaptation.
\newblock In {\em Advances in Neural Information Processing Systems (NIPS)},
  pages 3733--3742, 2017.

\bibitem{csurka2017comprehensive}
G.~Csurka.
\newblock A comprehensive survey on domain adaptation for visual applications.
\newblock pages 1--35. Springer, 2017.

\bibitem{deng2009imagenet}
J.~Deng, W.~Dong, R.~Socher, L.-J. Li, K.~Li, and L.~Fei-Fei.
\newblock Imagenet: A large-scale hierarchical image database.
\newblock In {\em Computer Vision and Pattern Recognition, 2009. CVPR 2009.
  IEEE Conference on}, pages 248--255. Ieee, 2009.

\bibitem{derpanis2008bhattacharyya}
K.~G. Derpanis.
\newblock The bhattacharyya measure.
\newblock {\em Mendeley Computer}, 1(4):1990--1992, 2008.

\bibitem{dezfouli15}
A.~Dezfouli and E.~V. Bonilla.
\newblock Scalable inference for {G}aussian process models with black-box
  likelihoods, 2015.
\newblock In Advances in Neural Information Processing Systems.

\bibitem{ganin2014unsupervised}
Y.~Ganin and V.~Lempitsky.
\newblock Unsupervised domain adaptation by backpropagation.
\newblock {\em International Conference on Machine Learning (ICML)}, 2015.

\bibitem{grl16}
Y.~Ganin, E.~Ustinova, H.~Ajakan, P.~Germain, H.~Larochelle, F.~Laviolette,
  M.~Marchand, and V.~Lempitsky.
\newblock Domain adversarial training of neural networks.
\newblock {\em Journal of Machine Learning Research}, 17(59):1--35, 2016.

\bibitem{ghifary2016deep}
M.~Ghifary, W.~B. Kleijn, M.~Zhang, D.~Balduzzi, and W.~Li.
\newblock Deep reconstruction-classification networks for unsupervised domain
  adaptation.
\newblock In {\em Euroupean Conference on Computer Vision (ECCV)}, pages
  597--613, 2016.

\bibitem{gholami2017punda}
B.~Gholami, V.~Pavlovic, et~al.
\newblock Punda: Probabilistic unsupervised domain adaptation for knowledge
  transfer across visual categories.
\newblock In {\em Proceedings of the IEEE International Conference on Computer
  Vision}, pages 3581--3590, 2017.

\bibitem{gong2013connecting}
B.~Gong, K.~Grauman, and F.~Sha.
\newblock Connecting the dots with landmarks: Discriminatively learning
  domain-invariant features for unsupervised domain adaptation.
\newblock In {\em International Conference on Machine Learning (ICML)}, pages
  222--230, 2013.

\bibitem{gan14}
I.~Goodfellow, J.~Pouget-Abadie, M.~Mirza, B.~Xu, D.~Warde-Farley, S.~Ozair,
  A.~Courville, and Y.~Bengio.
\newblock Generative adversarial nets, 2014.
\newblock In Advances in Neural Information Processing Systems.

\bibitem{gopalan2011domain}
R.~Gopalan, R.~Li, and R.~Chellappa.
\newblock Domain adaptation for object recognition: An unsupervised approach.
\newblock In {\em IEEE International Conference on Computer Vision (ICCV)},
  pages 999--1006, 2011.

\bibitem{ssem04}
Y.~Grandvalet and Y.~Bengio.
\newblock Semi-supervised learning by entropy minimization, 2004.
\newblock {I}n Proc. of Advances in Neural Information Processing Systems.

\bibitem{mmd}
A.~Gretton, K.~M. Borgwardt, M.~J. Rasch, B.~Schölkopf, and A.~Smola.
\newblock A kernel two-sample test.
\newblock {\em Journal of Machine Learning Research}, 13(1):723--773, 2012.

\bibitem{haeusser2017associative}
P.~Haeusser, T.~Frerix, A.~Mordvintsev, and D.~Cremers.
\newblock Associative domain adaptation.
\newblock In {\em International Conference on Computer Vision (ICCV)},
  volume~2, page~6, 2017.

\bibitem{he2016deep}
K.~He, X.~Zhang, S.~Ren, and J.~Sun.
\newblock Deep residual learning for image recognition.
\newblock In {\em Proceedings of the IEEE conference on computer vision and
  pattern recognition}, pages 770--778, 2016.

\bibitem{hu2018duplex}
L.~Hu, M.~Kan, S.~Shan, and X.~Chen.
\newblock Duplex generative adversarial network for unsupervised domain
  adaptation.
\newblock In {\em Proceedings of the IEEE Conference on Computer Vision and
  Pattern Recognition}, pages 1498--1507, 2018.

\bibitem{deep_kernel}
W.~Huang, D.~Zhao, F.~Sun, H.~Liu, and E.~Chang.
\newblock Scalable {G}aussian process regression using deep neural networks,
  2015.
\newblock Proceedings of the Twenty-Fourth International Joint Conference on
  Artificial Intelligence (IJCAI).

\bibitem{kan2015bi}
M.~Kan, S.~Shan, and X.~Chen.
\newblock Bi-shifting auto-encoder for unsupervised domain adaptation.
\newblock In {\em IEEE International Conference on Computer Vision (ICCV)},
  pages 3846--3854, 2015.

\bibitem{kingma2014adam}
D.~P. Kingma and J.~Ba.
\newblock Adam: A method for stochastic optimization.
\newblock {\em International Conference on Learning Representation (ICLR)},
  2015.

\bibitem{vae14}
D.~P. Kingma and M.~Welling.
\newblock Auto-encoding variational {B}ayes, 2014.
\newblock In Proceedings of the Second International Conference on Learning
  Representations, ICLR.

\bibitem{lecun1998gradient}
Y.~LeCun, L.~Bottou, Y.~Bengio, and P.~Haffner.
\newblock Gradient-based learning applied to document recognition.
\newblock {\em Proceedings of the IEEE}, 86(11):2278--2324, 1998.

\bibitem{lin2014microsoft}
T.-Y. Lin, M.~Maire, S.~Belongie, J.~Hays, P.~Perona, D.~Ramanan,
  P.~Doll{\'a}r, and C.~L. Zitnick.
\newblock Microsoft coco: Common objects in context.
\newblock In {\em European conference on computer vision}, pages 740--755.
  Springer, 2014.

\bibitem{liu2017unsupervised}
M.-Y. Liu, T.~Breuel, and J.~Kautz.
\newblock Unsupervised image-to-image translation networks.
\newblock In {\em Advances in Neural Information Processing Systems (NIPS)},
  pages 700--708, 2017.

\bibitem{NIPS2016_6544}
M.-Y. Liu and O.~Tuzel.
\newblock Coupled generative adversarial networks.
\newblock In D.~D. Lee, M.~Sugiyama, U.~V. Luxburg, I.~Guyon, and R.~Garnett,
  editors, {\em Advances in Neural Information Processing Systems (NIPS)},
  pages 469--477. Curran Associates, Inc., 2016.

\bibitem{long2015learning}
M.~Long, Y.~Cao, J.~Wang, and M.~I. Jordan.
\newblock Learning transferable features with deep adaptation networks.
\newblock {\em International Conference on Machine Learning (ICML)}, 2015.

\bibitem{long2014transfer}
M.~Long, J.~Wang, G.~Ding, J.~Sun, and P.~S. Yu.
\newblock Transfer joint matching for unsupervised domain adaptation.
\newblock In {\em IEEE Conference on Computer Vision and Pattern Recognition
  (CVPR)}, pages 1410--1417, 2014.

\bibitem{maaten2008visualizing}
L.~v.~d. Maaten and G.~Hinton.
\newblock Visualizing data using t-sne.
\newblock {\em Journal of Machine Learning Research}, 9(Nov):2579--2605, 2008.

\bibitem{mancini2018boosting}
M.~Mancini, L.~Porzi, S.~R. Bul{\`o}, B.~Caputo, and E.~Ricci.
\newblock Boosting domain adaptation by discovering latent domains.
\newblock {\em arXiv preprint arXiv:1805.01386}, 2018.

\bibitem{ming2015unsupervised}
T.~Ming Harry~Hsu, W.~Yu~Chen, C.-A. Hou, Y.-H. Hubert~Tsai, Y.-R. Yeh, and
  Y.-C. Frank~Wang.
\newblock Unsupervised domain adaptation with imbalanced cross-domain data.
\newblock In {\em IEEE International Conference on Computer Vision (ICCV)},
  pages 4121--4129, 2015.

\bibitem{moiseev2013evaluation}
B.~Moiseev, A.~Konev, A.~Chigorin, and A.~Konushin.
\newblock Evaluation of traffic sign recognition methods trained on
  synthetically generated data.
\newblock In {\em International Conference on Advanced Concepts for Intelligent
  Vision Systems}, pages 576--583. Springer, 2013.

\bibitem{motiian2017few}
S.~Motiian, Q.~Jones, S.~Iranmanesh, and G.~Doretto.
\newblock Few-shot adversarial domain adaptation.
\newblock In {\em Advances in Neural Information Processing Systems (NIPS)},
  pages 6673--6683, 2017.

\bibitem{murez2017image}
Z.~Murez, S.~Kolouri, D.~Kriegman, R.~Ramamoorthi, and K.~Kim.
\newblock Image to image translation for domain adaptation.
\newblock {\em arXiv preprint arXiv:1712.00479}, 13, 2017.

\bibitem{netzer2011reading}
Y.~Netzer, T.~Wang, A.~Coates, A.~Bissacco, B.~Wu, and A.~Y. Ng.
\newblock Reading digits in natural images with unsupervised feature learning.
\newblock In {\em NIPS workshop on deep learning and unsupervised feature
  learning}, volume 2011, page~5, 2011.

\bibitem{peng2017visda}
X.~Peng, B.~Usman, N.~Kaushik, J.~Hoffman, D.~Wang, and K.~Saenko.
\newblock Visda: The visual domain adaptation challenge.
\newblock {\em arXiv preprint arXiv:1710.06924}, 2017.

\bibitem{quinonero05}
J.~Qui{\~{n}}onero-Candela and C.~E. Rasmussen.
\newblock A unifying view of sparse approximate {G}aussian process regression.
\newblock {\em Journal of Machine Learning Research}, 6:1939--1959, 2005.

\bibitem{rf_fourier}
A.~Rahimi and B.~Recht.
\newblock Random features for large-scale kernel machines, 2008.
\newblock In Platt, J. C., Koller, D., Singer, Y., and Roweis, S. T. (eds.),
  Advances in Neural Information Processing Systems 20.

\bibitem{gpml_book}
C.~E. Rasmussen and C.~K.~I. Williams.
\newblock {\em Gaussian Processes for Machine Learning}.
\newblock The MIT Press, 2006.

\bibitem{rebuffi2017learning}
S.-A. Rebuffi, H.~Bilen, and A.~Vedaldi.
\newblock Learning multiple visual domains with residual adapters.
\newblock In {\em Advances in Neural Information Processing Systems (NIPS)},
  pages 506--516, 2017.

\bibitem{rozantsev2018residual}
A.~Rozantsev, M.~Salzmann, and P.~Fua.
\newblock Residual parameter transfer for deep domain adaptation.
\newblock In {\em Conference on Computer Vision and Pattern Recognition}, 2018.

\bibitem{saito2017asymmetric}
K.~Saito, Y.~Ushiku, and T.~Harada.
\newblock Asymmetric tri-training for unsupervised domain adaptation.
\newblock {\em International Conference on Machine Learning (ICML)}, 2017.

\bibitem{saito2018}
K.~Saito, K.~Watanabe, Y.~Ushiku, and T.~Harada.
\newblock Maximum classifier discrepancy for unsupervised domain adaptation.
\newblock {\em Computer Vision and Pattern Recognition}, 2018.

\bibitem{sankaranarayanan2017generate}
S.~Sankaranarayanan, Y.~Balaji, C.~D. Castillo, and R.~Chellappa.
\newblock Generate to adapt: Aligning domains using generative adversarial
  networks.
\newblock {\em ArXiv e-prints, abs/1704.01705}, 2017.

\bibitem{snelson06}
E.~Snelson and Z.~Ghahramani.
\newblock Sparse {G}aussian processes using pseudo-inputs, 2006.
\newblock In Advances in Neural Information Processing Systems.

\bibitem{stallkamp2011german}
J.~Stallkamp, M.~Schlipsing, J.~Salmen, and C.~Igel.
\newblock The german traffic sign recognition benchmark: a multi-class
  classification competition.
\newblock In {\em Neural Networks (IJCNN), The 2011 International Joint
  Conference on}, pages 1453--1460. IEEE, 2011.

\bibitem{sugiyama2008direct}
M.~Sugiyama, S.~Nakajima, H.~Kashima, P.~V. Buenau, and M.~Kawanabe.
\newblock Direct importance estimation with model selection and its application
  to covariate shift adaptation.
\newblock In {\em Advances in Neural Information Processing Systems (NIPS)},
  pages 1433--1440, 2008.

\bibitem{sun2016deep}
B.~Sun and K.~Saenko.
\newblock Deep coral: Correlation alignment for deep domain adaptation.
\newblock In {\em European Conference on Computer Vision (ECCV)}, pages
  443--450. Springer, 2016.

\bibitem{titsias09}
M.~K. Titsias.
\newblock Variational learning of inducing variables in sparse {G}aussian
  processes, 2009.
\newblock In Proceedings of the Twelfth International Conference on Artificial
  Intelligence and Statistics.

\bibitem{tzeng2017adversarial}
E.~Tzeng, J.~Hoffman, K.~Saenko, and T.~Darrell.
\newblock Adversarial discriminative domain adaptation.
\newblock In {\em IEEE Conference on Computer Vision and Pattern Recognition
  (CVPR)}, volume~1, page~4, 2017.

\bibitem{dom_conf}
E.~Tzeng, J.~Hoffman, N.~Zhang, K.~Saenko, and T.~Darrell.
\newblock Deep domain confusion: {M}aximizing for domain invariance, 2014.
\newblock arXiv:1412.3474.

\bibitem{vapnik_book98}
V.~Vapnik.
\newblock {\em Statistical Learning Theory}.
\newblock Wiley-Interscience, 1998.

\bibitem{wang2013max}
X.~Wang, B.~Wang, X.~Bai, W.~Liu, and Z.~Tu.
\newblock Max-margin multiple-instance dictionary learning.
\newblock In {\em International Conference on Machine Learning}, pages
  846--854, 2013.

\bibitem{dkl16}
A.~G. Wilson, Z.~Hu, R.~Salakhutdinov, and E.~P. Xing.
\newblock Deep kernel learning, 2016.
\newblock AI and Statistics (AISTATS).

\bibitem{Yan_2017_CVPR}
H.~Yan, Y.~Ding, P.~Li, Q.~Wang, Y.~Xu, and W.~Zuo.
\newblock Mind the class weight bias: Weighted maximum mean discrepancy for
  unsupervised domain adaptation.
\newblock In {\em IEEE Conference on Computer Vision and Pattern Recognition
  (CVPR)}, July 2017.

\bibitem{zellinger2017central}
W.~Zellinger, T.~Grubinger, E.~Lughofer, T.~Natschl{\"a}ger, and
  S.~Saminger-Platz.
\newblock Central moment discrepancy (cmd) for domain-invariant representation
  learning.
\newblock {\em International Conference on Learning Representation (ICLR)},
  2017.

\bibitem{Zhang_2017_CVPR}
J.~Zhang, W.~Li, and P.~Ogunbona.
\newblock Joint geometrical and statistical alignment for visual domain
  adaptation.
\newblock In {\em IEEE Conference on Computer Vision and Pattern Recognition
  (CVPR)}, July 2017.

\bibitem{zhang2015deep}
X.~Zhang, F.~X. Yu, S.-F. Chang, and S.~Wang.
\newblock Deep transfer network: Unsupervised domain adaptation.
\newblock {\em arXiv preprint arXiv:1503.00591}, 2015.

\bibitem{zhang2018aligning}
Z.~Zhang, M.~Wang, Y.~Huang, and A.~Nehorai.
\newblock Aligning infinite-dimensional covariance matrices in reproducing
  kernel hilbert spaces for domain adaptation.
\newblock In {\em Proceedings of the IEEE Conference on Computer Vision and
  Pattern Recognition}, pages 3437--3445, 2018.

\bibitem{zhu2017unpaired}
J.-Y. Zhu, T.~Park, P.~Isola, and A.~A. Efros.
\newblock Unpaired image-to-image translation using cycle-consistent
  adversarial networks.
\newblock {\em arXiv preprint}, 2017.

\bibitem{semisup_book}
X.~Zhu and A.~B. Goldberg.
\newblock {\em Introduction to semi-supervised learning}.
\newblock Morgan \& Claypool, 2009.

\end{thebibliography}

\end{document}